\documentclass[sigconf,nonacm]{acmart} %for arxiv

\AtBeginDocument{%
  \providecommand\BibTeX{{%
    \normalfont B\kern-0.5em{\scshape i\kern-0.25em b}\kern-0.8em\TeX}}}
    
\usepackage{mwe}
\usepackage{graphicx}
\usepackage{amsmath,amsfonts}
\usepackage{svg}
\usepackage{caption}
\usepackage{subcaption}
\usepackage{hyperref}
\usepackage{multirow}
\usepackage{tabularx}
\usepackage{colortbl}
\usepackage{xcolor}
\usepackage{booktabs} % for better table rules
\usepackage{textcomp}
\usepackage{booktabs}
\usepackage{rotating} % for rotated text
\usepackage{array} % for fixed width columns
\usepackage{adjustbox}
\usepackage{siunitx}
\usepackage{tikz}
\usepackage{subcaption}
\usepackage{threeparttable}
\usepackage{stfloats}
\usepackage{siunitx}
\usepackage{mathtools}
\newcommand{\rc}{\textcolor{red}{\raisebox{0.15ex}{\small$\bullet$}}\,}
\newcommand{\gc}{\textcolor{green!70!black}{\raisebox{0.15ex}{\small$\bullet$}}\,}
\renewcommand\footnotetextcopyrightpermission[1]{} % No ACM copyright
\acmConference[]{}{}{} % Clear conference info
\acmYear{} % Clear year
\acmISBN{} % Clear ISBN
\acmDOI{} % Clear DOI
\begin{document}

\title{Over-Squashing in GNNs and Causal Inference of Rewiring Strategies}

\author{Danial Saber}
\email{danial.saber@ontariotechu.ca}
\affiliation{%
  \institution{Ontario Tech University}
  \city{Oshawa}
  \state{Ontario}
  \country{Canada}
}

\author{Amirali Salehi-Abari}
\email{abari@ontariotechu.ca}
\affiliation{%
  \institution{Ontario Tech University}
  \city{Oshawa}
  \state{Ontario}
  \country{Canada}
}

\renewcommand{\shortauthors}{Danial Saber and Amirali Salehi-Abari} 

\begin{abstract}
Graph neural networks (GNNs) have exhibited state-of-the-art performance across wide-range of domains such as recommender systems, material design, and drug repurposing.  Yet message-passing GNNs suffer from over-squashing---exponential compression of long-range information from distant nodes---which limits expressivity. Rewiring techniques can ease this bottleneck; but their practical impacts are unclear due to the lack of a direct empirical over-squashing metric.
We propose a rigorous, topology-focused method for assessing over-squashing between node pairs using the decay rate of their mutual sensitivity. We then extend these pairwise assessments to four graph-level statistics (prevalence, intensity, variability, extremity). Coupling these metrics with a within-graph causal design, we quantify how rewiring strategies affect over-squashing on diverse graph- and node-classification benchmarks. Our extensive empirical analyses show that most graph classification datasets suffer from over-squashing (but to various extents), and rewiring effectively mitigates it---though the degree of mitigation, and its translation into performance gains, varies by dataset and method. We also found that over-squashing is less notable in node classification datasets, where rewiring often increases over-squashing, and performance variations are uncorrelated with over-squashing changes. These findings suggest that rewiring is most beneficial when over-squashing is both substantial and corrected with restraint---while overly aggressive rewiring, or rewiring applied to minimally over-squashed graphs, is unlikely to help and may even harm performance. Our plug-and-play diagnostic tool lets practitioners decide---before any training---whether rewiring is likely to pay off.

\end{abstract}

% \begin{CCSXML}
% <ccs2012>
%    <concept>
%     <concept_id>10010147.10010257.10010293.10010294</concept_id>
%        <concept_desc>Computing methodologies~Neural networks</concept_desc>
%        <concept_significance>500</concept_significance>
%        </concept>
%  </ccs2012>
% \end{CCSXML}

% \ccsdesc[500]{Computing methodologies~Neural networks}

\keywords{Graph Neural Networks, Over-Squashing, Rewiring}

\settopmatter{printfolios=true}
\maketitle

\section{Introduction}
Graph Neural Networks (GNNs) \cite{gori2005new,micheli2009neural,scarselli2008graph} have become a powerful learning framework for graph-structured data. Message-passing Neural Networks (MPNNs) \cite{gilmer2017neural}---a prominent subclass of GNNs--- iteratively aggregate messages from neighboring nodes at each layer, enabling information propagation across the graph through layer stacking. To enable interactions between distant nodes, deeper networks with more layers are often required \cite{barcelo2020logical}. However, as the number of layers increases, the receptive field of each node (i.e., the set of nodes that influence a node's representation through message passing) can expand rapidly, leading to excessive compression of information into fixed-size node representations. This phenomenon, known as \textit{over-squashing} \cite{alon2021on}, ultimately hampers effective information flow and learning.

As over-squashing is strongly connected with the topological properties of input graphs (e.g., commute time and effective resistance \cite{black2023understanding,di2023over}), most of its mitigation approaches are rewiring techniques \cite{topping2021understanding,alon2021on,nguyen2023revisiting,arnaiz2022diffwire,gasteiger2019diffusion}, which modify a graph’s connectivity to improve information flow between distant, weakly connected nodes. Despite its promise, the effectiveness of rewiring techniques remains challenging to assess due to the absence of a direct, empirical measure of over-squashing.
The Jacobian norm offers a formal foundation for measuring over-squashing, but it is computationally prohibitive in time and memory, and does not isolate the graph's topology effect on over-squashing due to its high dependency on the model's choices and parameters. Due to these limitations, effective resistance has emerged as a proxy \cite{black2023understanding,di2023over}, which offers relative insights---e.g., which of two node-pairs (or two graphs) are more susceptible to suffering over-squashing. However, this measurement lacks a clear threshold to identify (e.g., whether or not over-squashing occurs for a node pair or a graph) or quantify the extent of over-squashing. This ambiguity obscures the need or justification for rewiring. Even when rewiring improves performance, we still cannot tell whether the resulting effective resistance value remains harmful or whether the observed gains arise from reduced over-squashing rather than from other implicit structural benefits such as regularization or changes in graph smoothness.

To tackle these challenges, we propose a topology-focused measurement framework for over-squashing built upon a formal characterization of over-squashing---rather than being a proxy. We quantify pairwise over-squashing by modeling node-pair sensitivity exponentially decaying with the model depth (i.e., number of layers). This assumption mirrors the over-squashing theoretical definitions of Topping et al. \cite{topping2021understanding}, which show that sensitivity diminishes rapidly along long paths in over-squashed graphs. Using decay rates of node pairs as a direct and interpretable indicator of over-squashing, we derive graph-level over-squashing metrics and then leverage them in a causal inference framework to evaluate the rewiring effectiveness for over-squashing mitigation. This enables a rigorous evaluation of rewiring strategies on over-squashing across a diverse range of graph and node classification tasks.

We applied our measurement framework to address four key questions across node- and graph-classification tasks: \textit{(extent)} How much over-squashing does each dataset exhibit?; \textit{(mitigation)} How effectively do current rewiring methods reduce it?; \textit{(translation)} Do these reductions translate into performance gains?; and \textit{(responsiveness)} How responsive is each dataset to over-squashing mitigation? Our results show that most graph classification datasets suffer from substantial over-squashing (except Reddit-B), making rewiring a sensible intervention. Among the rewiring strategies, DIGL \cite{gasteiger2019diffusion} is the most effective in mitigating over-squashing, yet FoSR \cite{karhadkar2022fosr} and BORF \cite{nguyen2023revisiting} exhibit stronger correlations between over-squashing reduction and performance improvements (i.e., more effective in translation). Every graph dataset is responsive---over-squashing falls after rewiring---except Reddit-B, which is counter-responsive. 
%In node classification datasets, over-squashing is only pronounced in Chameleon. In others (e.g., Citeseer, Cora, Cornell, Texas, Wisconsin), rewiring often increases over-squashing levels, and performance changes are independent of it (i.e., no translation). Also, node datasets are mostly counter-responsive to the rewiring.
In most node-classification datasets, rewiring often increases over-squashing, and performance changes are independent of it (i.e., no translation). Also, node datasets are mostly counter-responsive to the rewiring. 
Our findings suggest that rewiring is most effective when over-squashing is a significant issue, as in most graph-classification datasets, and less justified when over-squashing is minimal (as in most node-classification datasets). Our plug-and-play diagnostic framework enables practitioners to quantify over-squashing and decide---before expending training cycles---whether rewiring is likely to pay off, thus focusing on the most relevant remedies for addressing over-squashing issues. 

\section{Preliminaries and Related Work}
\label{sec:pre}
We consider an undirected graph $G = (V, E)$ with $n$ nodes and $m$ edges, represented by its adjacency matrix $\mathbf{A} \in \mathbb{R}^{n\times n}$. To include self-loops, we define $\mathbf{\tilde{A}} = \mathbf{A} + \mathbf{I}$, with $\mathbf{I}\in\mathbb{R}^{n\times n}$ being the identity matrix. Each node $v$ has a $d$-dimensional feature vector $\mathbf{x}_v \in \mathbb{R}^d$. Message-Passing Neural Networks (MPNNs) propagate information through the graph by the $L$-stack of graph convolution layers (or message-passing layers), where $L$ represents the depth of the model. Each layer $\ell$ is composed of the aggregator  $\text{agg}^{(\ell)}$ function (e.g., mean)  and update $\text{up}^{(\ell)}$  function (e.g., MLP). Node $v$'s representation $\mathbf{h}_v^{(\ell)}$ at layer $\ell$ is updated by
\begin{equation}
    \mathbf{h}_v^{(\ell)} = \text{up}^{(\ell)}\left(\mathbf{h}_v^{(\ell - 1)}, \text{agg}^{(\ell)}\left(\{\mathbf{h}_u^{(\ell - 1)\\}:u \in N_v\}\right)\right),
\end{equation}
where $N_v = \{u \in V : (u,v) \in E\}$ denotes the 1-hop neighborhood of node $v$. The number of layers $L$ (i.e., model depth) determines how far information can flow across the graph, by defining the \textit{receptive field} for each node $v$---the set of nodes whose initial features (at layer 0) can influence $v$'s final representation $\mathbf{h}_v^{(L)}$. As each layer propagates information one hop further, the receptive field grows with the depth of the model. 

\vskip 0.1cm
\noindent \textbf{Over-Squashing.}
When a task relies on long-range information exchange between distant node pairs, effective information propagation requires the model depth $L$ to be at least as large as the geodesic distance between nodes, allowing them to fall within each other's receptive fields. However, in most real-world graphs, receptive fields grow exponentially with the number of layers, forcing MPNNs to compress increasingly large sets of node features into fixed-width node embeddings. This excessive compression leads to information loss and reduces the model's expressivity, a phenomenon known as \textit{over-squashing} \cite{alon2021on}. The over-squashing of information can be understood by assessing the sensitivity of node $v$'s representation after $\ell$ layers of message passing to node $u$'s input feature $\mathbf{h}_u^{(0)}$ through the \textit{absolute Jacobian’s norm} \cite{topping2021understanding}:\footnote{Some use the term \emph{influence} for the same Jacobian‑based norm quantity; we adopt \emph{sensitivity} throughout for consistency.}  
\begin{equation}
\mathcal{J}_\ell(v, u)=\|\partial \mathbf{h}_v^{(\ell)}/ \partial \mathbf{h}_u^{(0)}\|.    
\label{eq:abs_jacob}
\end{equation}
Of special interest for assessing over-squashing is \textit{normalized Jacobian’s norm}  
\begin{equation}
    \mathcal{\tilde{J}}_\ell(v, u) = \frac{\mathcal{J}_\ell(v,u)}{\sum_{k}\mathcal{J}_\ell(v,k)},
    \label{eq:norm_jacob}
\end{equation} 
which measures relative sensitivity \cite{topping2021understanding,xu2018representation}---the sensitivity of node $v$'s feature at layer $\ell$ to node $u$'s initial feature, relative to $v$'s sensitivity to all nodes. Without this normalization, the model might overestimate a node’s sensitivity based solely on its absolute Jacobian norm.
A small $\mathcal{\tilde{J}}_\ell(v, u)$ indicates that node $v$ is negligibly sensitive to node $u$, signaling over-squashing. In severe cases---e.g., tree-like graphs \cite{topping2021understanding}---both $\mathcal{J}_\ell(v, u)$ and $\mathcal{\tilde{J}}_\ell(v, u)$ decay exponentially with $\ell$, causing sensitivity to vanish. This vanishing sensitivity reflects the progressive suppression of messages from $u$ at $v$, a signature of over-squashing.

Our over-squashing measures build upon the theoretical characterization of $\mathcal{\tilde{J}}_\ell(v, u)$, which links the decay of pairwise sensitivity in node embeddings to the number of GNN layers and the graph’s topology.

\phantomsection
\label{sec:over_squashing_measurement}
\vskip 0.1cm
\noindent \textbf{Over-Squashing Measurement.}
The Jacobian norm (and its variants) is a principled measure of over-squashing, but has practical shortcomings. (i) It is \textit{computationally prohibitive}: for $n$ nodes with feature dimension $d$, the full Jacobian is an $(nd) \times (nd)$ matrix, requiring $O(n^2d^2)$ memory and time. (ii) It is \textit{parameter-dependent}, varying with weight updates and model-specific hyperparameters (e.g., hidden size). 
(iii) It fails to isolate the graph's topological effects, being highly dependent on model choices and parameters. 

To focus more directly on graph topology, recent work resorts to measuring over-squashing through the lens of effective resistance \cite{di2023over,black2023understanding}, capturing key topological properties such as commute time and the Cheeger constant. Node pairs with high effective resistance are more susceptible to over-squashing \cite{di2023over,black2023understanding}, and a graph’s total effective resistance serves as a global proxy for over-squashing. However, effective resistance has key limitations: (a) it only allows relative comparisons---offering no threshold for when, or how severely, over-squashing occurs; (b) Though related to over-squashing, the effective resistance is not derived from its formal characterizations (e.g., $\mathcal{\tilde{J}}_\ell(v, u)$), leaving uncertainty about whether a given pair is truly over-squashed. To avoid these shortcomings, we approximate the \emph{relative Jacobian norm} directly, yielding a measure that is both topology-centered and largely model-agnostic without the heavy computational cost of full Jacobian computation.

\vskip 0.1cm
\noindent \textbf{Rewiring.}
Rewiring---the primary mitigation for over-squashing \cite{alon2021on}---transforms the input graph $G = (V, E)$ by keeping the node set $V$ unchanged, but altering the edge set $E$. The goal of this alteration is to enhance information propagation across the graph. A class of rewiring methods optimizes spatial connectivity by adding more edges between nodes to reduce large distances. Examples include introducing well-connected virtual nodes \cite{cai2023connection,southern2024understanding}, leveraging higher-order structures \cite{bodnar2021weisfeilera,bodnar2021weisfeilerb} for message-passing, fully connecting the graph in the last GNN layer \cite{alon2021on}, and propagating information between the nodes by connecting them within a certain distance and different layers \cite{abboud2022shortest,abu2019mixhop,wang2020multi,nikolentzos2020k,gasteiger2019diffusion,gabrielsson2022rewiring,barbero2023locality,gutteridge2023drew}. An extreme form of spatial rewiring is in Graph Transformers, where all nodes are connected to others through attention-based edges \cite{ying2021transformers,kreuzer2021rethinking,rampavsek2022recipe}.

Other rewiring approaches optimize other graph-theoretical properties (such as curvature \cite{topping2021understanding,nguyen2023revisiting}, spectral gap \cite{karhadkar2022fosr}, effective resistance \cite{black2023understanding}, and personalized PageRank \cite{gasteiger2019diffusion}) to alleviate topological bottlenecks. SDRF \cite{topping2021understanding} adds edges in low-curvature regions, while BORF \cite{nguyen2023revisiting} adds edges to minimally curved regions and prunes the highly curved edges. FoSR \cite{karhadkar2022fosr} iteratively maximizes the spectral gap to improve information flow. GTR \cite{black2023understanding} repeatedly adds edges to the graph to minimize the graph's total effective resistance. DIGL \cite{gasteiger2019diffusion}---a diffusion-based rewiring scheme---computes a generalized graph diffusion (e.g., personalized PageRank or heat kernel) from the adjacency matrix, followed by sparsification.

While rewiring aims to mitigate over-squashing, its true impact remains unclear; performance gains may arise from reduced over-squashing or other confounding factors, such as implicit regularization or altered graph smoothness. To disentangle these effects, we propose a measurement framework that leverages causal inference to evaluate the impact of rewiring interventions. This approach enables a systematic evaluation of the influence of established rewiring techniques (e.g, SDRF \cite{topping2021understanding}, FoSR \cite{karhadkar2022fosr}, DIGL \cite{gasteiger2019diffusion}, GTR \cite{black2023understanding}, and BORF \cite{nguyen2023revisiting}) on over-squashing. These methods are standard rewiring baselines in prior studies \cite{barbero2023locality,karhadkar2022fosr,nguyen2023revisiting, black2023understanding}.

\section{Measurement and Causality Framework}
We propose a theoretically-founded, topology-focused method to measure pairwise over-squashing for pairs of nodes in graphs and extend it to quantify over-squashing at the graph level. With our graph-level measurement, we assess the causal effect of rewiring techniques in mitigating over-squashing.

\subsection{Pairwise Over-Squashing Measurement}
Our goal is to derive a pairwise over-squashing measure between node pairs in a graph that is (i) computed once per graph, (ii) aligned with the relative Jacobian norm as a foundation for measuring over-squashing, (iii) focused on graph topology, (iv) dependent only on model depth as a contributing factor,\footnote{Model depth is necessary for any over-squashing measurement as the definition of over-squashing is based on it: the progressive compression of information as the model depth increases (i.e., the number of message-passing layers grows).} and (v) theoretically-founded on the rigorous definition of over-squashing. We achieve (i--iv) by introducing approximations to relative Jacobian norms, and (v) by considering the exponential decay rate.

\vskip 1mm
\noindent \textbf{Approximation to Normalized Jacobian Norm}.
To quantify over-squashing, we focus on the relative Jacobian norm $\mathcal{\tilde{J}}_\ell(v, u)$ as it
captures node $v$'s sensitivity to node $u$ as a fraction of its total sensitivity to all nodes, addressing the limitation of the absolute norm $\mathcal{{J}}_\ell(v, u)$, which does not consider the overall amount of information $v$ receives. Computing $\mathcal{{\tilde{J}}}_\ell(v, u)$ is prohibitively expensive due to the cost of Jacobian computation (see Section \ref{sec:over_squashing_measurement} for details), and its requirement for recomputation with the changes of the model parameters and hyperparameters (e.g., weights, hidden-dimensionality, etc.). Additionally, these dependencies blur the topological causes of over-squashing with its model-level effects. To address all these computational and dependency issues, we introduce approximations to normalized Jacobian norms.

We observe that the normalized Jacobian norm is entirely formulated by absolute Jacobian terms (see Equation \ref{eq:norm_jacob}). Thus, by approximating absolute Jacobian norms, we can readily achieve an approximation for the normalized Jacobian norm.

\begin{proposition}[Approximation of the Normalized Jacobian Norm]\label{prop:rel_Jacobian}
Let $\mathbf{\tilde A}= \mathbf A + \mathbf I$ be the adjacency matrix of an undirected graph augmented with self-loops, and assume a \emph{linear} message-passing GNN. Then, for any pair of nodes $u,v$ and layer depth $\ell \ge 0$, the \emph{normalized} Jacobian norm can be written as
\begin{equation}
    \mathcal{\tilde{J}}_\ell(v, u) = \frac{{\mathbf{\tilde{A}}}^{\ell}_{uv}}{\sum_{k}{\mathbf{\tilde{A}}}^{\ell}_{kv}},
    \label{eq:rel-approx}
\end{equation}
where $\mathbf{\tilde{A}}_{uv}^\ell$ is the $(u,v)$-entry of the matrix $\mathbf{\tilde{A}}$ raised to the power $\ell$.
\end{proposition}

\noindent \emph{Remark.} Self-loops guarantee reachability for every choice of $\ell$: e.g., in a dyad (two nodes joined by an edge), walks of even length between nodes vanish without self-loops, but $\mathbf{\tilde A}$ ensures non-zero counts for all $\ell$. 

The proof of the proposition is in Appendix \ref{app:relativeapprox}. In practice, GNNs typically include nonlinearities (e.g., ReLU), which yield computation of nontrivial Jacobians that require recursive application of the chain rule. Moreover, certain paths in the computational graph may become inactive (e.g., due to ReLU zeroing gradients), making exact computation intractable. Thus, equality no longer holds for nonlinear MPNNs. However, linear MPNNs are empirically competitive and theoretically well-founded \cite{wu2019simplifying,saber2024scalable,frasca2020sign,louis2023simplifying}, and omitting nonlinearity helps remove model-specific factors (see the proof of Proposition \ref{prop:rel_Jacobian}), enabling a sensitivity measure that reflects only the graph structure and the depth.

We approximate the nonlinear normalized Jacobian’s norm using its simplified single-computation form in Eq.~\ref{eq:rel-approx}, and henceforth denote this approximation as $\mathcal{\tilde{J}}_\ell(v,u)$. This approximation satisfies our design criteria outlined earlier: it is once-pre-computed (criterion i), derived from normalized Jacobian norms under some simplification assumptions (i.e., removing non-linearities) 
(criterion ii), and only dependent on graph topology $\mathbf{\tilde{A}}$ and the layer depth $\ell$, remaining mainly topology-focused and model-agnostic (criteria iii and iv)..

\vskip 1mm
\noindent \textbf{Exponential Decay Rate as an Over-Squashing Indicator}.
A key signature of over-squashing is the rapid decay of sensitivity (e.g., normalized Jacobian norms) with increasing model depth $\ell$. For rigor, we model this decay as exponential, similar to Di Giovanni et al. \cite{di2023over}, and consistent with theoretical observations in tree-like graphs \cite{topping2021understanding}, where both $\mathcal{J}_\ell(v, u)$ and $\mathcal{\tilde{J}}_\ell(v, u)$ diminish exponentially with $\ell$, leading to vanishing sensitivity: 
\begin{equation}
    \mathcal{\tilde{J}}_\ell(v, u) = N_0e^{-k_{vu}\ell},
\end{equation}
where $N_0 = \mathcal{\tilde J}_0(v, u)$ is the initial sensitivity ($\ell = 0$), and $k_{vu}$ is the decay rate specific to the pair $(v,u)$.\footnote{The decay rates $k_{vu}$ and $k_{uv}$ need not be equal because the number of length-$\ell$ walks that reach the target node can differ for $v$ and $u$. 
For instance, in a star graph with center $v$ and a leaf $u$, we have $\mathcal{\tilde J}_1(v,u) \neq \mathcal{\tilde J}_1(u,v)$ since $(\sum_k \tilde A^1_{kv} \neq \sum_k \tilde A^1_{ku})$.}
 A positive $k_{vu}$ indicates over-squashing, with larger values reflecting stronger decay. Taking the natural logarithm linearizes this relationship:

\begin{equation}
    \ln \mathcal{\tilde{J}}_\ell(v, u) = \ln N_0 - k_{vu}\ell.
\end{equation}
To estimate $k_{vu}$, we fit a linear regression model of $\ln \mathcal{\tilde{J}}_\ell(v, u)$ against $\ell$, where the slope corresponds to $-k_{vu}$. A negative slope (i.e., positive $k_{vu}$) confirms exponential decay, with the magnitude of $k_{vu}$ reflecting the severity of over-squashing.\footnote{For specific pairwise analyses, statistics such as coefficient of determination $R^2$ and $p$-value assess model fit and the statistical significance of the decay trend. However, since our focus is on graph-level over-squashing, we aggregate pairwise decay rates into graph-level metrics and evaluate statistical significance within our causal framework.} Following Di Giovanni et al. \cite{di2023over}, we change $\ell$ in the interval $[\text{D}, 2 \text{D} - 1]$, where $D$ is the graph diameter, ensuring reachability for any pair of nodes.

\subsection{Graph-Level Over-Squashing Measurement}
To derive a graph-level assessment, we summarize the distribution of positive decay rates using four statistics:
\begin{itemize}
    \item \textbf{Prevalence} is the fraction of node pairs with positive decay rates ($k_{vu} > 0$). It reflects the \emph{spread} of over-squashing across the graph.  
    \item \textbf{Intensity} is the average of all positive decay rates, indicating the typical \emph{strength} of over-squashing among affected node pairs. 
    \item \textbf{Variability} is the standard deviation of positive decay rates, measuring the \emph{consistency or disparity} in over-squashing strength across node pairs. 
    \item \textbf{Extremity} is the largest observed positive decay rate in the graph, capturing the \emph{worst-case} over-squashing instance.
\end{itemize}
For datasets involving multiple graphs, we compute dataset-level summaries by averaging each metric over all graphs. For ease of interpretation, we sometimes map each graph-level statistic into three ordinal categories. Intensity and Extremity categorized as \textit{weak} $(< 0.13)$, \textit{moderate} $(0.13\text{--}0.23)$, \textit{strong} $(> 0.23)$, based on corresponding \textit{pairwise sensitivity half-lives} of $\geq5$, 3--5, and $<3$ layers, respectively (i.e., the number of layers needed for sensitivity to halve). %The pairwise sensitivity half-life represents the number of layers required for the pairwise sensitivity to decay to half of its initial value.
Under the same thresholds, variability is classified as \textit{low} $(< 0.13)$, \textit{moderate} $(0.13\text{--}0.23)$, \textit{high} $(> 0.23)$. Prevalence is grouped as \textit{small} $(< 25\%)$, \textit{moderate} $(25-50\%)$, \textit{large} $(>50\%)$; these cut-points align with intuitive quartile boundaries: fewer than one quarter of node pairs indicates sparse over-squashing, 25–50\%  a moderate regime, and more than 50\% an affected majority.

% Together, these four metrics provide a comprehensive view of over-squashing in terms of its prevalence, intensity, variability, and extremes. For datasets involving multiple graphs, we compute dataset-level summaries by averaging each metric over all graphs.

\subsection{Causal Estimation of Rewiring Effects}
To assess the impact of any specific rewiring technique $\mathcal{R}$ on over-squashing, we leverage our graph-level over-squashing metrics in a causal inference framework, where rewiring $\mathcal{R}$ serves as the treatment $T$, and graph-level over-squashing measurements (i.e., prevalence, intensity, variability, and extremity) represent the outcome.

The treatment variable $T$ is a binary variable indicating whether the intervention (i.e., rewiring $\mathcal{R}$) is applied: $ T=1$ for \textit{treated} (rewired) graph  $\mathcal{R}(G)$, and $T=0$ for the  \textit{controlled} (original) graph $G$.
The outcome (i.e., a graph-level measurement of prevalence, intensity, variability, or extremity) is observed under two conditions: $Y_{\mathcal{M}}(G)$ for the original graph (control condition), and $Y_{\mathcal{M}}(\mathcal{R}(G))$ for the rewired graph (treated condition), where $\mathcal{M}$ can be any graph measurement for over-squashing. The goal is to estimate the causal effect of $T$ on outcome $Y$ for each graph $G(V, E)$, treating each graph as an individual unit in the analysis. This within-unit design (comparing each graph before and after rewiring) helps isolate the effect of rewiring from confounding factors tied to graph topology (e.g., number of nodes). To ensure valid causal attribution of differences in over-squashing metrics to rewiring, we adopt standard causal inference assumptions: SUTVA, Positivity, Exchangeability, and Consistency. A detailed explanation of these assumptions and their relevance to our setup is provided in Appendix \ref{app:causality}.

We assess how a rewiring $\mathcal{R}$ influences the over-squashing measurement $\mathcal{M}$ for a graph $G$ through \textit{Individual Treatment Effect (ITE)}:
\begin{equation}
    \text{ITE}_{\mathcal{M}}(G,\mathcal{R}) = Y_{\mathcal{M}}(\mathcal{R}(G)) - Y_{\mathcal{M}}(G).
\end{equation}
For graph classification tasks with a dataset of $N$ graphs $\mathcal{D} = \{G_i\}$, we measure the \textit{Average Treatment Effect (ATE)} to capture the overall impact of rewiring $\mathcal{R}$ across the dataset:
\begin{equation}
    \text{ATE}_{\mathcal{M}}(\mathcal{D}, \mathcal{R}) = \frac{1}{N}\sum_{i=1}^N \text{ITE}_{\mathcal{M}}(G_i, \mathcal{R}).
\end{equation}
For each dataset, we analyze all four graph-level measurements to evaluate the effect of rewiring $\mathcal{R}$ on over-squashing prevalence, intensity, variability, and extremity within the graph. For example, a negative ATE on prevalence indicates that rewiring reduces the number of over-squashed node pairs; a negative ATE on intensity reflects a decrease in the average severity of over-squashing; and a negative ATE on extremity suggests that the most severe cases of over-squashing have been mitigated.

\vskip 0.1cm
\noindent \textbf{Statistical Significance of Treatment Effects.} Estimating decay rates via linear regression involves uncertainty, as some pairwise decay rates $k_{vu}$ may arise from random noise rather than genuine over-squashing effects. This uncertainty propagates to graph-level metrics and treatment effects (ITE/ATE).\footnote{Despite potential noise, we noticed most pairs exhibit statistically significant decay rates. The Law of Large Numbers mitigates random errors by aggregating hundreds or thousands of pairs.}
For graph classification tasks, we test the significance of $\text{ATE}$ using a two-tailed $t$-test, justified by the Central Limit Theorem \cite{dudley2014uniform} for $N > 30$. To control family-wise error rate (FWER) across four tests (each for one measurement), we apply the Bonferroni correction \cite{weisstein2004bonferroni}, adjusting the significance threshold $\alpha$ by dividing it by the number of tests. For node classification, we assess ITE significance, shifting the unit of analysis from graphs to node pairs to leverage their variability. We test $\text{ITE}$ on prevalence with McNemar’s test \cite{mcnemar1947note}, a non-parametric method for paired binary data, and $\text{ITE}$ on intensity using a paired $t$-test. 

\section{Experiments}
We first measure the over-squashing levels across various datasets. Then, using our causal framework, we evaluate the effectiveness of rewiring techniques in mitigating over-squashing on graph and node classification benchmarks.

\begin{table}[t]
\caption{Statistics of graph-classification datasets, averaged over all graphs in each dataset. Color coding shows over-squashing extent: \colorbox{green!20}{weak/low/small}, \colorbox{yellow!30}{moderate}, and \colorbox{red!20}{strong/high/large}.}
\label{tab:graphdatasetstats}
\centering
\small
\setlength{\tabcolsep}{1mm}
\renewcommand{\arraystretch}{1.1}
\begin{tabularx}{\linewidth}{@{}l@{\hspace{0.6em}}l
  @{\hspace{0.5em}}>{\centering\arraybackslash}X
  @{\hspace{0.5em}}>{\centering\arraybackslash}X
  @{\hspace{0.5em}}>{\centering\arraybackslash}X
  @{\hspace{0.5em}}>{\centering\arraybackslash}X
  @{\hspace{0.5em}}>{\centering\arraybackslash}X
  @{\hspace{0.5em}}>{\centering\arraybackslash}X@{}}
\toprule
 & Statistic                      
 & \multicolumn{3}{c}{\text{Bioinformatics}} 
 & \multicolumn{3}{c}{\text{Social Networks}} \\

 &  
 & Mutag & Proteins & Enzymes 
 & IMDB & Collab & Reddit \\
\cmidrule(lr){1-2}\cmidrule(lr){3-5}\cmidrule(lr){6-8}
\multirow{4}{*}{\rotatebox[origin=c]{90}{\tiny \text{Topology}}} 
 & \#Graphs     & $188$ & $1109$ & $600$ & $1000$ & $5000$ & $2000$ \\
 &  Nodes       & $18$  & $39$   & $33$  & $20$   & $74$   & $430$  \\
 &  Edges       & $28$  & $92$   & $78$  & $106$  & $2494$ & $712$  \\
 &  Diameter    & $8.21$& $11.56$& $10.89$& $1.86$& $1.86$ & $9$    \\
 &  Components  & $1.00$& $1.07$ & $1.24$ & $1.00$& $1.00$ & $2.48$ \\
\cmidrule{2-8}
\multirow{4}{*}{\rotatebox[origin=c]{90}{\tiny \text{Over-Squashing}}} 
    &  Prevalence    
        & \cellcolor{red!20}{$5.93\mathrm{e}\text{-}1$}
        & \cellcolor{red!20}{$5.97\mathrm{e}\text{-}1$}
        & \cellcolor{red!20}{$6.03\mathrm{e}\text{-}1$}
        & \cellcolor{red!20}{$6.28\mathrm{e}\text{-}1$}
        & \cellcolor{red!20}{$5.57\mathrm{e}\text{-}1$}
        & \cellcolor{yellow!20}{$4.72\mathrm{e}\text{-}1$} \\
    &  Intensity     
        & \cellcolor{green!20}{$1.09\mathrm{e}\text{-}1$}
        & \cellcolor{yellow!20}{$1.37\mathrm{e}\text{-}1$}
        & \cellcolor{yellow!20}{$1.30\mathrm{e}\text{-}1$}
        & \cellcolor{red!20}{$3.12\mathrm{e}\text{-}1$}
        & \cellcolor{red!20}{$2.56\mathrm{e}\text{-}1$}
        & \cellcolor{green!20}{$1.96\mathrm{e}\text{-}2$} \\
    &  Variability   
        & \cellcolor{green!20}{$1.06\mathrm{e}\text{-}1$}
        & \cellcolor{yellow!20}{$1.34\mathrm{e}\text{-}1$}
        & \cellcolor{yellow!20}{$1.31\mathrm{e}\text{-}1$}
        & \cellcolor{yellow!20}{$1.57\mathrm{e}\text{-}1$}
        & \cellcolor{yellow!20}{$1.93\mathrm{e}\text{-}1$}
        & \cellcolor{green!20}{$1.88\mathrm{e}\text{-}2$} \\
    &  Extremity     
        & \cellcolor{red!20}{$4.54\mathrm{e}\text{-}1$}
        & \cellcolor{red!20}{$5.71\mathrm{e}\text{-}1$}
        & \cellcolor{red!20}{$5.96\mathrm{e}\text{-}1$}
        & \cellcolor{red!20}{$5.49\mathrm{e}\text{-}1$}
        & \cellcolor{red!20}{$9.10\mathrm{e}\text{-}1$}
        & \cellcolor{yellow!20}{$1.35\mathrm{e}\text{-}1$} \\
\bottomrule
\end{tabularx}
\end{table}

\subsection{Methodology and Experimental Setup}
We discuss our experimental methodology, including empirical research questions, datasets, rewiring baselines, hyperparameters, measurements, and statistical tests.  

\vskip 2mm
\noindent \textbf{Research Questions and their Importance.}
In our experiments, using our measurement and causal inference framework, we address four key questions for graph and node-level tasks:
\begin{itemize}
    \item (\textbf{Q1}) \textit{How do over-squashing measurements (i.e., prevalence, intensity, variability, and extremity) vary across datasets? Which datasets are inherently most or least susceptible under each measurement?} This question identifies which datasets are inherently more or less prone to over-squashing, guiding benchmark selection for over-squashing research and the necessity of mitigation strategies. It also informs whether over-squashing trends are dataset-, or domain-specific (e.g., social vs. biological networks). Each metric captures distinct structural bottlenecks: For example, extremity highlights the most severe local cases, while intensity reflects global compression across the graph. 
    \item (\textbf{Q2}) \textit{What are the treatment effects of each rewiring method across datasets? Which rewiring strategy most (or least) effectively reduces over-squashing measurements?} This question quantifies the treatment effects of rewiring strategies and enables their comparative ``effective'' ranking. 
    \item (\textbf{Q3}) \textit{How do treatment effects correlate with performance gains for each rewiring strategy over all datasets?} This question evaluates whether reducing over-squashing translates into improved generalization. By assessing the correlation between treatment effects and performance gains (i.e., the change in predictive performance before and after rewiring), we can distinguish rewiring methods that improve performance by mitigating over-squashing from those whose gains arise from other factors. It also reveals strategies that reduce over-squashing without boosting performance, or even worsen it.
    \item (\textbf{Q4}) \textit{Which datasets are most responsive to rewiring---that is, show the largest relative reductions  (treatment effect divided by the pre-treatment value)---and which are most resistant?} Answering this question sheds light on the inherent difficulty of reducing over-squashing across different graph structures, datasets, or domains.
\end{itemize}

\vskip 2mm
\noindent \textbf{Datasets.}
We study node and graph classification datasets commonly employed in over-squashing and rewiring research \cite{alon2021on,topping2021understanding,karhadkar2022fosr,nguyen2023revisiting,gasteiger2019diffusion,southern2024understanding}.
Graph classification datasets are from the TUDataset benchmark \cite{morris2020tudataset}: three \textit{bioinformatics} datasets of Mutag, Enzymes, and Proteins, and three \textit{social network} datasets of IMDB-B, Collab, and Reddit-B. We summarize their statistics in Table \ref{tab:graphdatasetstats}. For node classification, we evaluate six datasets: Cora, Citeseer \cite{yang2016revisiting}, Texas, Cornell, Wisconsin \cite{pei2020geom}, and Chameleon \cite{rozemberczki2021multi}, analyzing only their largest connected component (see Table \ref{tab:nodedatasetstats} for statistics).

\vskip 2mm
\noindent \textbf{Rewiring Baselines (Subjects).}
We examine the effectiveness of five rewiring methods commonly used in mitigating over-squashing%\cite{karhadkar2022fosr,barbero2023locality,banerjee2022oversquashing,nguyen2023revisiting,choi2024panda}
: FoSR \cite{karhadkar2022fosr}, DIGL \cite{gasteiger2019diffusion}, SDRF \cite{topping2021understanding}, GTR \cite{black2023understanding}\footnote{Since prior work evaluated GTR only on graph classification, we also restrict our study of it to that task.}, and BORF \cite{nguyen2023revisiting}% (discussed in Section \ref{sec:pre})
.\footnote{For future work, one can easily extend our experiments to dynamic and training-time rewiring methods. In this work, to avoid hyperparameter tuning or ad-hoc design choices, we focused on the widely used methods, whose reported results are based on a shared experimental setup.} 
\vskip 2mm
\noindent \textbf{Hyperparameters.}
Our measurement framework requires no hyperparameter tuning, enhancing its practicality. Its only parameter is the message-passing depth $\ell$, varied from the graph’s diameter to twice the diameter, following \cite{di2023over}. Rewiring methods, however, have their own hyperparameters---e.g., the number of iterations for FoSR and SDRF, the number of edges added or removed per iteration for BORF, the number of edges added in GTR, and the sparsification threshold for DIGL)---all tuned for specific GNN architecture (e.g., GCN, GIN, etc.) used for performance evaluation. To control for this dependency, we assess each rewiring method using all performance-optimal hyperparameter configurations identified in prior work for each GNN architecture. For graph classification, each rewiring method is evaluated using four sets of optimal hyperparameters---one for each of GCN, GIN, R-GCN, and R-GIN---whereas there are two sets of configurations for node classification: GCN and GIN. This approach prevents bias from relying on a single configuration and ensures fair comparisons across different GNN architectures. When a rewiring method is not evaluated with a particular GNN (e.g., BORF with R-GCN/R-GIN or DIGL with node-level GIN), we omit that combination rather than performing ad-hoc retuning.

\vskip 2mm
\noindent \textbf{Measurements.}
To address \textit{Q1–Q4}, we apply our over-squashing measurement framework. For \textit{Q1}, we measure over-squashing metrics (e.g., prevalence, intensity, etc.) on the original graphs (see Tables \ref{tab:graphdatasetstats} and \ref{tab:nodedatasetstats}). For \textit{Q2}--\textit{Q4}, we compute the individual treatment effect $\text{ITE}_\mathcal{M}(G, \mathcal{R)}$ for node classification and average treatment effect $\text{ATE}_\mathcal{M}(\mathcal{D}, \mathcal{R)}$ for graph classification based on changes in over-squashing after each rewiring treatment. Each rewiring method is evaluated using its original performance-optimal, architecture-specific hyperparameters, yielding one ATE/ITE per (method, architecture) pair. 
To control for architectural dependency and avoid bias from any single configuration, we report \textit{aggregated ATE/ITE} values by averaging across all relevant configurations (see Tables \ref{tab:graphresults} and \ref{tab:noderesults}). 
%Because each rewiring method is tied to multiple GNN-specific configurations, we report an accumulated ATE/ITE, computed by averaging the values across all such configurations (Tables \ref{tab:graphresults} and \ref{tab:noderesults}). This aggregation is crucial---it controls for the dependence of treatment effects on model architecture and avoids biased conclusions based on a single configuration.
Statistical significance is assessed at $\alpha=0.05$ with Bonferroni correction \cite{weisstein2004bonferroni}.
Performance gains (i.e., the change in task performance before and after rewiring) are directly taken from the original papers under the same replicated hyperparameter settings.\footnote{The code is available at https://github.com/Danial-sb/Over-Squashing-Measurement.}

\begin{table}[t]
\caption{Statistics of node-classification datasets. Color coding shows
over-squashing extent: \colorbox{green!20}{weak/low/small}, \colorbox{yellow!30}{moderate}, and \colorbox{red!20}{strong/high/large}.}
\label{tab:nodedatasetstats}
\centering
\small
\setlength{\tabcolsep}{1mm}
\renewcommand{\arraystretch}{1.1}
\begin{tabularx}{1.0\columnwidth}{@{}l@{\hspace{0.6em}} l @{\hspace{1em}}   >{\centering\arraybackslash}X@{}
  >{\centering\arraybackslash}X@{}
  >{\centering\arraybackslash}X@{}
  >{\centering\arraybackslash}X@{}
  >{\centering\arraybackslash}X@{}
  >{\centering\arraybackslash}X
  @{}}
\toprule
 & Statistic & Cornell & Texas & Wisconsin & Cora & Citeseer & Chamel. \\
\midrule
\multirow{3}{*}{\rotatebox[origin=c]{90}{\tiny \text{Topology}}} 
    & \#Nodes    & $140$  & $135$  & $184$   & $2485$  & $2120$  & $832$ \\
 & \#Edges    & $219$  & $251$  & $362$   & $5096$  & $3679$  & $12355$ \\
 & Diameter   & $8$    & $8$    & $8$     & $19$    & $28$    & $11$ \\
\cmidrule{2-8}
\multirow{4}{*}{\rotatebox[origin=c]{90}{\tiny \text{Over-Squashing}}} 
    & Prevalence
        & \cellcolor{red!20}{$5.47\mathrm{e}\text{-}1$}
        & \cellcolor{red!20}{$5.02\mathrm{e}\text{-}1$}
        & \cellcolor{red!20}{$5.46\mathrm{e}\text{-}1$}
        & \cellcolor{green!20}{$1.52\mathrm{e}\text{-}2$}
        & \cellcolor{green!20}{$1.84\mathrm{e}\text{-}3$}
        & \cellcolor{green!20}{$2.03\mathrm{e}\text{-}1$} \\
    & Intensity
        & \cellcolor{green!20}{$8.99\mathrm{e}\text{-}3$}
        & \cellcolor{green!20}{$6.20\mathrm{e}\text{-}3$}
        & \cellcolor{green!20}{$7.95\mathrm{e}\text{-}3$}
        & \cellcolor{green!20}{$3.63\mathrm{e}\text{-}2$}
        & \cellcolor{green!20}{$3.09\mathrm{e}\text{-}4$}
        & \cellcolor{yellow!30}{$1.42\mathrm{e}\text{-}1$} \\
    & Variability
        & \cellcolor{green!20}{$8.02\mathrm{e}\text{-}3$}
        & \cellcolor{green!20}{$5.72\mathrm{e}\text{-}3$}
        & \cellcolor{green!20}{$8.01\mathrm{e}\text{-}3$}
        & \cellcolor{green!20}{$2.59\mathrm{e}\text{-}2$}
        & \cellcolor{green!20}{$1.76\mathrm{e}\text{-}3$}
        & \cellcolor{green!20}{$8.37\mathrm{e}\text{-}2$} \\
    & Extremity
        & \cellcolor{green!20}{$1.14\mathrm{e}\text{-}1$}
        & \cellcolor{green!20}{$8.04\mathrm{e}\text{-}2$}
        & \cellcolor{green!20}{$1.09\mathrm{e}\text{-}1$}
        & \cellcolor{yellow!30}{$2.04\mathrm{e}\text{-}1$}
        & \cellcolor{green!20}{$1.84\mathrm{e}\text{-}2$}
        & \cellcolor{red!20}{$3.94\mathrm{e}\text{-}1$} \\
\bottomrule
\end{tabularx}
\end{table}

\newcolumntype{Y}{>{\centering\arraybackslash}X}
\newcolumntype{B}{>{\centering\arraybackslash}m{1em}}
\begin{table*}[ht]
\centering
\caption{Treatment effects (ATEs) for graph classification, averaged over each GNN baseline's hyperparameter configuration (GCN, GIN, R-GCN, and R-GIN). For every dataset–metric cell, background shading highlights the \colorbox{green!20}{best} and \colorbox{red!20}{worst} rewiring method. A green-filled circle (\gc) marks desirable negative ATEs, while a red-filled circle (\rc) marks undesirable positive ones. Gain is the percentage change in classification accuracy after rewiring. $\dagger$ marks not statistically significant results. Avg ATE reports the average treatment effect of each rewiring method across all datasets.}
\label{tab:graphresults}
\begin{tabularx}{\textwidth}{@{}l l *{4}{Y@{\hspace{-0.6em}}B}Y@{}}\toprule
\textbf{Rew.} & \textbf{Dataset} &
\multicolumn{8}{c}{\textbf{Average Treatment Effect}} &
\textbf{Gain (\%)}\\
\cmidrule(lr){3-10}
& &
\textbf{Prevalence} & &
\textbf{Intensity}  & &
\textbf{Variability} & &
\textbf{Extremity} &
\\
\midrule
\multirow{7}{*}{FoSR}
& Mutag      
  & $-2.3\mathrm{e}\text{-}2 \pm 6.3\mathrm{e}\text{-}3$ & \gc
  & $-1.5\mathrm{e}\text{-}2 \pm 5.5\mathrm{e}\text{-}3$ & \gc
  & $-2.7\mathrm{e}\text{-}2 \pm 9.3\mathrm{e}\text{-}3$ & \gc
  & $-3.9\mathrm{e}\text{-}2 \pm 2.1\mathrm{e}\text{-}2$ & \gc
  & $\phantom{-}6.6 \pm 6.5$ \\

& Proteins   
  & $-4.0\mathrm{e}\text{-}2 \pm 2.8\mathrm{e}\text{-}2$ & \gc
  & $-3.1\mathrm{e}\text{-}2 \pm 1.3\mathrm{e}\text{-}2$ & \gc
  & $-3.9\mathrm{e}\text{-}2 \pm 1.3\mathrm{e}\text{-}2$ & \gc
  & $-7.5\mathrm{e}\text{-}2 \pm 3.7\mathrm{e}\text{-}2$ & \gc
  & $\phantom{-}3.8 \pm 0.9$ \\

& Enzymes    
  & $-1.6\mathrm{e}\text{-}2 \pm 2.3\mathrm{e}\text{-}2$ & \gc
  & $-2.2\mathrm{e}\text{-}2 \pm 1.7\mathrm{e}\text{-}2$ & \gc
  & $-3.7\mathrm{e}\text{-}2 \pm 2.2\mathrm{e}\text{-}2$ & \gc
  & $-7.7\mathrm{e}\text{-}2 \pm 7.0\mathrm{e}\text{-}2$ & \gc
  & $\phantom{-}1.6 \pm 6.0$ \\

& IMDB\text{-}B 
  & \cellcolor{red!20}$-1.2\mathrm{e}\text{-}2 \pm 5.1\mathrm{e}\text{-}3$ & \gc
  & $-1.4\mathrm{e}\text{-}1 \pm 4.8\mathrm{e}\text{-}2$ & \gc
  & $\ \ -2.2\mathrm{e}\text{-}2 \pm 1.4\mathrm{e}\text{-}2^\dagger$ & \gc
  & $-1.4\mathrm{e}\text{-}2 \pm 2.0\mathrm{e}\text{-}0$ & \gc
  & $\phantom{-}4.1 \pm 6.6$ \\

& Collab     
  & $\ \ \, 8.9\mathrm{e}\text{-}3 \pm 1.0\mathrm{e}\text{-}3$ & \rc
  & $-1.2\mathrm{e}\text{-}2 \pm 9.7\mathrm{e}\text{-}3$ & \gc
  & $-8.0\mathrm{e}\text{-}3 \pm 8.9\mathrm{e}\text{-}3$ & \gc
  & $-1.5\mathrm{e}\text{-}2 \pm 2.7\mathrm{e}\text{-}3$ & \gc
  & $\phantom{-}\ \ 9.6 \pm 18.2$ \\

& Reddit\text{-}B 
  & $\ \ \ \ 5.4\mathrm{e}\text{-}4 \pm 1.2\mathrm{e}\text{-}4^\dagger$ & \rc
  & $\ \ \, 5.3\mathrm{e}\text{-}3 \pm 5.6\mathrm{e}\text{-}3$ & \rc
  & $\ \ \, 3.8\mathrm{e}\text{-}3 \pm 3.1\mathrm{e}\text{-}3$ & \rc
  & $\ \ \, 2.5\mathrm{e}\text{-}2 \pm 2.0\mathrm{e}\text{-}2$ & \rc
  & $\ \ \phantom{-}7.8 \pm 12.7$ \\
\cmidrule{2-11}
& Avg ATE   
  & $-1.4\mathrm{e}\text{-}2 \pm 1.7\mathrm{e}\text{-}2$ & \gc
  & $-3.6\mathrm{e}\text{-}2 \pm 5.2\mathrm{e}\text{-}2$ & \gc
  & $-2.2\mathrm{e}\text{-}2 \pm 1.7\mathrm{e}\text{-}2$ & \gc
  & $-3.3\mathrm{e}\text{-}2 \pm 3.9\mathrm{e}\text{-}2$ & \gc
  & $\phantom{-}5.6 \pm 2.9$ \\
\midrule
\multirow{7}{*}{DIGL}
& Mutag      
  & \cellcolor{green!20}$-5.0\mathrm{e}\text{-}1 \pm 1.9\mathrm{e}\text{-}1$ & \gc
  & \cellcolor{green!20}$-1.0\mathrm{e}\text{-}1 \pm 1.3\mathrm{e}\text{-}2$ & \gc
  & \cellcolor{green!20}$-9.9\mathrm{e}\text{-}2 \pm 1.3\mathrm{e}\text{-}2$ & \gc
  & \cellcolor{green!20}$-4.2\mathrm{e}\text{-}1 \pm 6.6\mathrm{e}\text{-}2$ & \gc
  & $\phantom{-} 0.7 \pm 3.0$ \\

& Proteins   
  & \cellcolor{green!20}$-3.4\mathrm{e}\text{-}1 \pm 1.1\mathrm{e}\text{-}1$ & \gc
  & \cellcolor{green!20}$-9.1\mathrm{e}\text{-}2 \pm 3.2\mathrm{e}\text{-}2$ & \gc
  & \cellcolor{green!20}$-8.5\mathrm{e}\text{-}2 \pm 3.2\mathrm{e}\text{-}2$ & \gc
  & \cellcolor{green!20}$-3.3\mathrm{e}\text{-}1 \pm 1.5\mathrm{e}\text{-}1$ & \gc
  & $ -0.2 \pm 0.9$ \\

& Enzymes    
  & \cellcolor{green!20}$-2.8\mathrm{e}\text{-}1 \pm 1.7\mathrm{e}\text{-}1$ & \gc
  & \cellcolor{green!20}$-6.7\mathrm{e}\text{-}2 \pm 4.6\mathrm{e}\text{-}2$ & \gc
  & \cellcolor{green!20}$-6.4\mathrm{e}\text{-}2 \pm 4.8\mathrm{e}\text{-}2$ & \gc
  & \cellcolor{green!20}$-2.7\mathrm{e}\text{-}1 \pm 2.3\mathrm{e}\text{-}1$ & \gc
  & $\phantom{-}0.0 \pm 1.4$ \\

& IMDB\text{-}B 
  & \cellcolor{green!20}$\! \! -6.3\mathrm{e}\text{-}1 \pm 0.0\mathrm{e}0$ & \gc
  & \cellcolor{green!20}$\! \! -3.1\mathrm{e}\text{-}1 \pm 0.0\mathrm{e}0$ & \gc
  & $\! \! -1.6\mathrm{e}\text{-}1 \pm 0.0\mathrm{e}0$ & \gc
  & \cellcolor{green!20}$\! \! -5.5\mathrm{e}\text{-}1 \pm 0.0\mathrm{e}0$ & \gc
  & $-2.9 \pm 3.3$ \\

& Collab     
  & \cellcolor{green!20}$-5.4\mathrm{e}\text{-}1 \pm 1.5\mathrm{e}\text{-}2$ & \gc
  & \cellcolor{green!20}$-2.6\mathrm{e}\text{-}1 \pm 5.8\mathrm{e}\text{-}4$ & \gc
  & \cellcolor{green!20}$-1.9\mathrm{e}\text{-}1 \pm 1.1\mathrm{e}\text{-}3$ & \gc
  & \cellcolor{green!20}$-9.0\mathrm{e}\text{-}1 \pm 1.3\mathrm{e}\text{-}2$ & \gc
  & $\! \!-18.2 \pm 1.1$ \\

& Reddit\text{-}B 
  & \cellcolor{green!20}$-9.4\mathrm{e}\text{-}2 \pm 5.9\mathrm{e}\text{-}2$ & \gc
  & \cellcolor{red!20}$\  \ \, 5.7\mathrm{e}\text{-}2 \pm 1.2\mathrm{e}\text{-}2$ & \rc
  & \cellcolor{red!20}$\ \ \, 6.2\mathrm{e}\text{-}2 \pm 1.1\mathrm{e}\text{-}2$ & \rc
  & \cellcolor{red!20}$\ \ \, 6.3\mathrm{e}\text{-}1 \pm 9.0\mathrm{e}\text{-}2$ & \rc
  & $\! \!-13.3 \pm 3.6$ \\
\cmidrule{2-11}
& Avg ATE   
  & \cellcolor{green!20}$-4.0\mathrm{e}\text{-}1 \pm 1.8\mathrm{e}\text{-}1$ & \gc
  & \cellcolor{green!20}$-1.3\mathrm{e}\text{-}1 \pm 1.2\mathrm{e}\text{-}1$ &  \gc
  & $-8.9\mathrm{e}\text{-}2 \pm 8.8\mathrm{e}\text{-}2$ & \gc
  & \cellcolor{green!20}$-3.1\mathrm{e}\text{-}1 \pm 4.7\mathrm{e}\text{-}1$ & \gc
  & $\ -5.6 \pm 7.4$ \\
\midrule
\multirow{7}{*}{SDRF}
& Mutag      
  & $\! \! -1.0\mathrm{e}\text{-}2 \pm 0.0\mathrm{e}0$ & \gc
  & $\ 2.3\mathrm{e}\text{-}3 \pm 0.0\mathrm{e}0$ & \rc
  & $\ 2.9\mathrm{e}\text{-}3 \pm 0.0\mathrm{e}0$ & \rc
  & $\ 2.4\mathrm{e}\text{-}2 \pm 0.0\mathrm{e}0$ & \rc
  & $\! \ \ -0.5 \pm 1.4$ \\

& Proteins   
  & $-2.3\mathrm{e}\text{-}2 \pm 6.1\mathrm{e}\text{-}3$ & \gc
  & \cellcolor{red!20}$\ \, -3.9\mathrm{e}\text{-}4 \pm 1.6\mathrm{e}\text{-}4^\dagger$ & \gc
  & \cellcolor{red!20}$\ \ \, 3.6\mathrm{e}\text{-}3 \pm 2.0\mathrm{e}\text{-}3$ & \rc
  & \cellcolor{red!20}$\ \ \, 1.2\mathrm{e}\text{-}2 \pm 3.8\mathrm{e}\text{-}3$ & \rc
  & $\! \ \ -0.3 \pm 0.5$ \\

& Enzymes    
  & $\! \! -1.3\mathrm{e}\text{-}2 \pm 0.0\mathrm{e}0$ & \gc
  & $\, -3.1\mathrm{e}\text{-}4 \pm 0.0\mathrm{e}0^\dagger$ & \gc
  & \cellcolor{red!20}$\! \! -2.4\mathrm{e}\text{-}3 \pm 0.0\mathrm{e}0$ & \gc
  & \cellcolor{red!20}$\ \ \ 1.3\mathrm{e}\text{-}2 \pm 0.0\mathrm{e}0^\dagger$ & \rc
  & $\ \ \ \, 2.0 \pm 2.0$ \\

& IMDB\text{-}B 
  & $-3.6\mathrm{e}\text{-}2 \pm 2.6\mathrm{e}\text{-}2$ & \gc
  & $ -4.4\mathrm{e}\text{-}2 \pm 2.8\mathrm{e}\text{-}2$ & \gc
  & \cellcolor{red!20}$\ \ \ \ \, 1.9\mathrm{e}\text{-}2 \pm 8.7\mathrm{e}\text{-}3^\dagger$ & \rc
  & \cellcolor{red!20}$\ \ \, 1.3\mathrm{e}\text{-}1 \pm 2.3\mathrm{e}\text{-}2$ & \rc
  & $\ \ \ \, 1.0 \pm 1.9$ \\

& Collab     
  & $\ \ \, 5.5\mathrm{e}\text{-}3 \pm 1.2\mathrm{e}\text{-}3$ & \rc
  & $-1.7\mathrm{e}\text{-}2 \pm 9.8\mathrm{e}\text{-}3$ & \gc
  & $-5.6\mathrm{e}\text{-}3 \pm 3.4\mathrm{e}\text{-}3$ & \gc
  & $ -3.7\mathrm{e}\text{-}2 \pm 1.6\mathrm{e}\text{-}2$ & \gc
  & $\ \ \ \ \ \, 8.8 \pm 17.2$ \\

& Reddit\text{-}B 
  & $\ \ -3.1\mathrm{e}\text{-}3 \pm 2.9\mathrm{e}\text{-}3^\dagger$ & \gc
  & \cellcolor{green!20}$ \ \ -1.1\mathrm{e}\text{-}4 \pm 1.1\mathrm{e}\text{-}3^\dagger$ & \gc
  & \cellcolor{green!20}$\ \ -6.5\mathrm{e}\text{-}4 \pm 7.7\mathrm{e}\text{-}4^\dagger$ & \gc
  & \cellcolor{green!20}$\ \, -2.9\mathrm{e}\text{-}4 \pm 2.7\mathrm{e}\text{-}3^\dagger$ & \gc
  & $\ \, -2.5 \pm 4.3$ \\
\cmidrule{2-11}
& Avg ATE   
  & $-1.3\mathrm{e}\text{-}2 \pm 1.5\mathrm{e}\text{-}2$ & \gc
  & $\ \! -9.9\mathrm{e}\text{-}3 \pm 1.8\mathrm{e}\text{-}2$ & \gc
  & \cellcolor{red!20}$\ \ \ 2.8\mathrm{e}\text{-}3 \pm 8.6\mathrm{e}\text{-}3$ & \rc
  & \cellcolor{red!20}$\ \ \, 2.4\mathrm{e}\text{-}2 \pm 5.6\mathrm{e}\text{-}2$ & \rc
  & $\ \ \ \ 1.4 \pm 3.6$ \\
\midrule
\multirow{7}{*}{GTR}
& Mutag      
  & \cellcolor{red!20}$-1.4\mathrm{e}\text{-}3 \pm 4.6\mathrm{e}\text{-}2$ & \gc
  & \cellcolor{red!20}$\ \ \ 8.9\mathrm{e}\text{-}2 \pm 3.6\mathrm{e}\text{-}2$ & \rc
  & \cellcolor{red!20}$\ \ \ 1.2\mathrm{e}\text{-}2 \pm 2.2\mathrm{e}\text{-}2$ & \rc
  & \cellcolor{red!20}$\ \ \ 5.4\mathrm{e}\text{-}2 \pm 6.4\mathrm{e}\text{-}2$ & \rc
  & $\ \ \ \ 6.5 \pm 7.1$ \\

& Proteins   
  & $-7.6\mathrm{e}\text{-}3 \pm 1.8\mathrm{e}\text{-}2$ & \gc
  & $\, -2.9\mathrm{e}\text{-}3 \pm 6.6\mathrm{e}\text{-}3$ & \gc
  & $\, -4.0\mathrm{e}\text{-}2 \pm 7.4\mathrm{e}\text{-}3$ & \gc
  & $\, -9.4\mathrm{e}\text{-}2 \pm 3.2\mathrm{e}\text{-}2$ & \gc
  & $\ \ \ \ 3.8 \pm 2.2$ \\

& Enzymes    
  & $\ \ \, 1.1\mathrm{e}\text{-}2 \pm 1.3\mathrm{e}\text{-}2$ & \rc
  & \cellcolor{red!20}$\ \ \, 1.3\mathrm{e}\text{-}2 \pm 4.5\mathrm{e}\text{-}3$ & \rc
  & $\ -3.2\mathrm{e}\text{-}2 \pm 4.7\mathrm{e}\text{-}3$ & \gc
  & $\ -8.0\mathrm{e}\text{-}2 \pm 2.0\mathrm{e}\text{-}2$ & \gc
  & $\ \ \ \ 5.1 \pm 7.9$ \\

& IMDB\text{-}B 
  & $-2.3\mathrm{e}\text{-}2 \pm 6.3\mathrm{e}\text{-}3$ & \gc
  & \cellcolor{red!20}$ -1.5\mathrm{e}\text{-}2 \pm 5.5\mathrm{e}\text{-}3$ & \gc
  & $\ -2.7\mathrm{e}\text{-}2 \pm 9.4\mathrm{e}\text{-}3$ & \gc
  & $\ -3.9\mathrm{e}\text{-}2 \pm 2.1\mathrm{e}\text{-}2$ & \gc
  & $\ \ \ \ 4.7 \pm 6.9$ \\

& Collab     
  & \cellcolor{red!20}$\ \ \, 1.5\mathrm{e}\text{-}2 \pm 1.8\mathrm{e}\text{-}3$ & \rc
  & \cellcolor{red!20}$\ \ \, 9.0\mathrm{e}\text{-}4 \pm 1.3\mathrm{e}\text{-}3$ & \rc
  & \cellcolor{red!20}$\ -8.8\mathrm{e}\text{-}4 \pm 2.3\mathrm{e}\text{-}4$ & \gc
  & \cellcolor{red!20}$\ \ \ \, 1.6\mathrm{e}\text{-}2 \pm 1.0\mathrm{e}\text{-}2$ & \rc
  & $\ \ \ \ 0.4 \pm 1.1$ \\

& Reddit\text{-}B 
  & \cellcolor{red!20}$\ \ \ \! 1.9\mathrm{e}\text{-}2 \pm 5.0\mathrm{e}\text{-}4$ & \rc
  & $\ \ \ \! \, 1.4\mathrm{e}\text{-}2 \pm 3.0\mathrm{e}\text{-}3$ & \rc
  & $\ \ \ \, 9.2\mathrm{e}\text{-}3 \pm 1.4\mathrm{e}\text{-}3$ & \rc
  & $\ \ \ \, 4.5\mathrm{e}\text{-}2 \pm 1.0\mathrm{e}\text{-}2$ & \rc
  & $\ \ \ \ \ \ 8.4 \pm 14.6$ \\
\cmidrule{2-11}
& Avg ATE   
  & \cellcolor{red!20}$\, \ \, 2.2\mathrm{e}\text{-}3 \pm 1.5\mathrm{e}\text{-}2$ & \rc
  & \cellcolor{red!20}$\ \ \, 1.7\mathrm{e}\text{-}2 \pm 3.4\mathrm{e}\text{-}2$ & \rc
  & $\, \, -1.3\mathrm{e}\text{-}2 \pm 2.1\mathrm{e}\text{-}2$     & \gc
  & $\, \, -1.6\mathrm{e}\text{-}2 \pm 5.8\mathrm{e}\text{-}2$     & \gc
  & $\ \ \ \ 4.8 \pm 2.5$ \\
\midrule
\multirow{5}{*}{BORF}
& Mutag     
  & $-8.0\mathrm{e}\text{-}2 \pm 1.2\mathrm{e}\text{-}2$ & \gc
  & $\ \! -2.8\mathrm{e}\text{-}2 \pm 2.5\mathrm{e}\text{-}2$ & \gc
  & $\ -2.6\mathrm{e}\text{-}3 \pm 3.3\mathrm{e}\text{-}3$ & \gc
  & $\, \, -1.3\mathrm{e}\text{-}1 \pm 1.3\mathrm{e}\text{-}1$ & \gc
  & $\ \ \ \ 2.5 \pm 1.8$ \\

& Proteins   
  & \cellcolor{red!20}$\ \ 1.5\mathrm{e}\text{-}2 \pm 6.3\mathrm{e}\text{-}3$ & \rc
  & $\ \! -6.4\mathrm{e}\text{-}3 \pm 2.8\mathrm{e}\text{-}3$ & \gc
  & $\ -2.4\mathrm{e}\text{-}2 \pm 1.1\mathrm{e}\text{-}2$ & \gc
  & $\, \, -1.3\mathrm{e}\text{-}1 \pm 3.1\mathrm{e}\text{-}2$ & \gc
  & $\ \ \ \ 0.4 \pm 0.1$ \\

& Enzymes    
  & \cellcolor{red!20}$\ \ \, 2.1\mathrm{e}\text{-}2 \pm 2.5\mathrm{e}\text{-}2$ & \rc
  & $\ \! -3.5\mathrm{e}\text{-}2 \pm 1.3\mathrm{e}\text{-}3$ & \gc
  & $\ -2.9\mathrm{e}\text{-}2 \pm 1.4\mathrm{e}\text{-}4$ & \gc
  & $\, \, -1.4\mathrm{e}\text{-}1 \pm 1.8\mathrm{e}\text{-}2$ & \gc
  & $\ \ \ \ 1.0 \pm 1.1$ \\

& IMDB\text{-}B 
  & $\! \! -5.1\mathrm{e}\text{-}2 \pm 0.0\mathrm{e}0$ & \gc
  & $\!\!-7.6\mathrm{e}\text{-}2 \pm 0.0\mathrm{e}0$ & \gc
  & \cellcolor{green!20}$\! \, -3.7\mathrm{e}\text{-}1 \pm 0.0\mathrm{e}0$ & \gc
  & $\, \, -2.1\mathrm{e}\text{-}1 \pm 9.9\mathrm{e}\text{-}2$ & \gc
  & $\ \ \ \ 0.9 \pm 1.0$ \\
\cmidrule{2-11}
& Avg ATE   
  & $-2.4\mathrm{e}\text{-}2 \pm 4.5\mathrm{e}\text{-}2$ & \gc
  & $\, \! -3.6\mathrm{e}\text{-}2 \pm 2.9\mathrm{e}\text{-}2$ & \gc
  & \cellcolor{green!20}$\ -1.1\mathrm{e}\text{-}1 \pm 1.8\mathrm{e}\text{-}1$ & \gc
  & $\, \, -1.5\mathrm{e}\text{-}1 \pm 3.9\mathrm{e}\text{-}2$ & \gc
  & $\ \ \ \ 1.2 \pm 0.9$ \\
\bottomrule
\end{tabularx}
\end{table*}

\subsection{Results on Graph Classification Tasks}
We report our results answering Q1--Q4 for graph classification. 

\vskip 0.03cm
\noindent \textbf{Dataset Over-Squashing Levels (Q1)}: Table \ref{tab:graphdatasetstats} shows that over-squashing prevalence is relatively consistent (55\%--62\%) across most datasets, except for Reddit-B, with a lower prevalence of 47\%.
For other measures, the bioinformatic datasets have low intensity (0.10--0.13), low variability (0.10--0.13), and high extremity (0.45--0.59). However, social network datasets exhibit more severe over-squashing: IMDB-B and Collab have the highest over-squashing intensities (0.31 and 0.25, respectively) over all datasets, with Collab showing the highest variability (0.19) and extremity (0.91). Reddit-B stands out as an outlier with all metrics an order of magnitude lower (intensity: 0.02, variability: 0.02, extremity: 0.13), confirming it is far less affected. Overall, social network datasets exhibit stronger over-squashing measurements---particularly in intensity and extremity---than bioinformatics datasets, with Collab being maximally over-squashed.

% \vskip 0.03cm
% \noindent \textbf{Rewiring Effectiveness (Q2)}:
% Figure \ref{fig:graphscatter} shows the ATE values for each rewiring technique obtained on GNN baselines' hyperparameter configurations. DIGL emerges as the most effective strategy for reducing over-squashing in graph classification tasks (see also Table \ref{tab:graphresults}), achieving up to a $67\%$ reduction in prevalence (at IMDB-B) and consistently lowering intensity, variability, and the extremity of over-squashed pairs more than the other methods. In contrast, the remaining strategies---based on curvature-driven rewiring---show similar, yet less substantial, treatment effects, underscoring their methodological similarity and contrast with DIGL.

\vskip 0.03cm
\noindent \textbf{Rewiring Effectiveness (Q2)}:
% Figure \ref{fig:graphscatter} shows the ATE for each rewiring method across datasets and GNN-specific hyperparameter configurations. 
Table~\ref{tab:graphresults} reports rewiring method's ATE across graph-classification datasets and shows that rewiring generally reduces over-squashing (green markers \gc dominate red markers \rc).
FoSR reduces all over-squashing metrics across datasets with a few exceptions on Collab (for prevalence) and Reddit-B (for all metrics). It lowers prevalence by up to $-0.04$ (Proteins), intensity by $-0.14$ (IMDB-B), variability by $-0.04$ (Proteins), and extremity by $-0.08$ (Enzymes). DIGL consistently mitigates over-squashing metrics in all datasets except Reddit-B, lowering prevalence by $28$–$63\%$, intensity by $0.067$–$0.31$, variability by $0.064$–$0.19$, and extremity by $0.27$–$0.90$. On Reddit-B, however, intensity ($+0.057$), variability ($+0.062$), and extremity ($+0.63$) increase, suggesting that aggressive rewiring on disconnected graphs can introduce new bottlenecks.
SDRF and GTR offer the weakest and most inconsistent effects. SDRF slightly reduces prevalence and intensity in most datasets by up to $-0.036$ (in IMDB-B for prevalence), while GTR often increases all metrics by up to $+0.089$ (in Mutag for intensity). Variability increases in three datasets for SDRF (Mutag, Proteins, IMDB-B) and two for GTR (Mutag, Reddit-B). Extremity worsens in four datasets under SDRF and in three under GTR.
% Prevalence decreases in most datasets (from $-0.003$ in Reddit-B to $-0.036$ in IMDB-B) except in Collab. Intensity shows near-zero reductions except in Mutag with an increment. Variability increases in three datasets (Mutag, Proteins, IMDB-B) and slightly decreases in the others. Extremity worsens in four cases and improves in only two.
BORF demonstrates reductions in most cases, with strong gains in reducing extremity (up to $-0.21$ on IMDB-B) and consistently lowering variability (from $-0.0026$ to $-0.37$) and intensity. However, it increases over-squashing prevalence in Proteins and Enzymes, while decreasing it in Mutag and IMDB-B.

Treatment-effect rankings show DIGL as the strongest mitigator in almost all cases, except four (three on Reddit-B%, where SDRF ranks first
). %FoSR yields the worst effect only for prevalence on IMDB-B, while BORF is least effective at reducing prevalence in Proteins and Enzymes, but performs best in mitigating over-squashing variability on IMDB-B. 
SDRF and GTR are usually the least effective, with SDRF being weakest in variability and extremity (on three datasets each) and GTR being worst for prevalence (on three datasets) and intensity (on four datasets).

Averaged over datasets (Avg. ATE in Table~\ref{tab:graphresults}), DIGL is most effective ($-40\%$ prevalence, $-0.13$ intensity, and $-0.31$ extremity). For variability, its effect is also close to the most effective strategy (BORF with $-0.11$).
SDRF and GTR are the least effective, with GTR showing adverse effects on prevalence and intensity, and SDRF performing worst in terms of variability and extremity.
Overall, aggressive densification (e.g., DIGL) alleviates over-squashing more effectively than surgical or sparsity-preserving rewiring (e.g., FoSR, SDRF, GTR, BORF).

\begin{figure*}[t]
    \centering
    \begin{subfigure}[b]{0.4\textwidth}
        \centering
        \includegraphics[width=\textwidth]{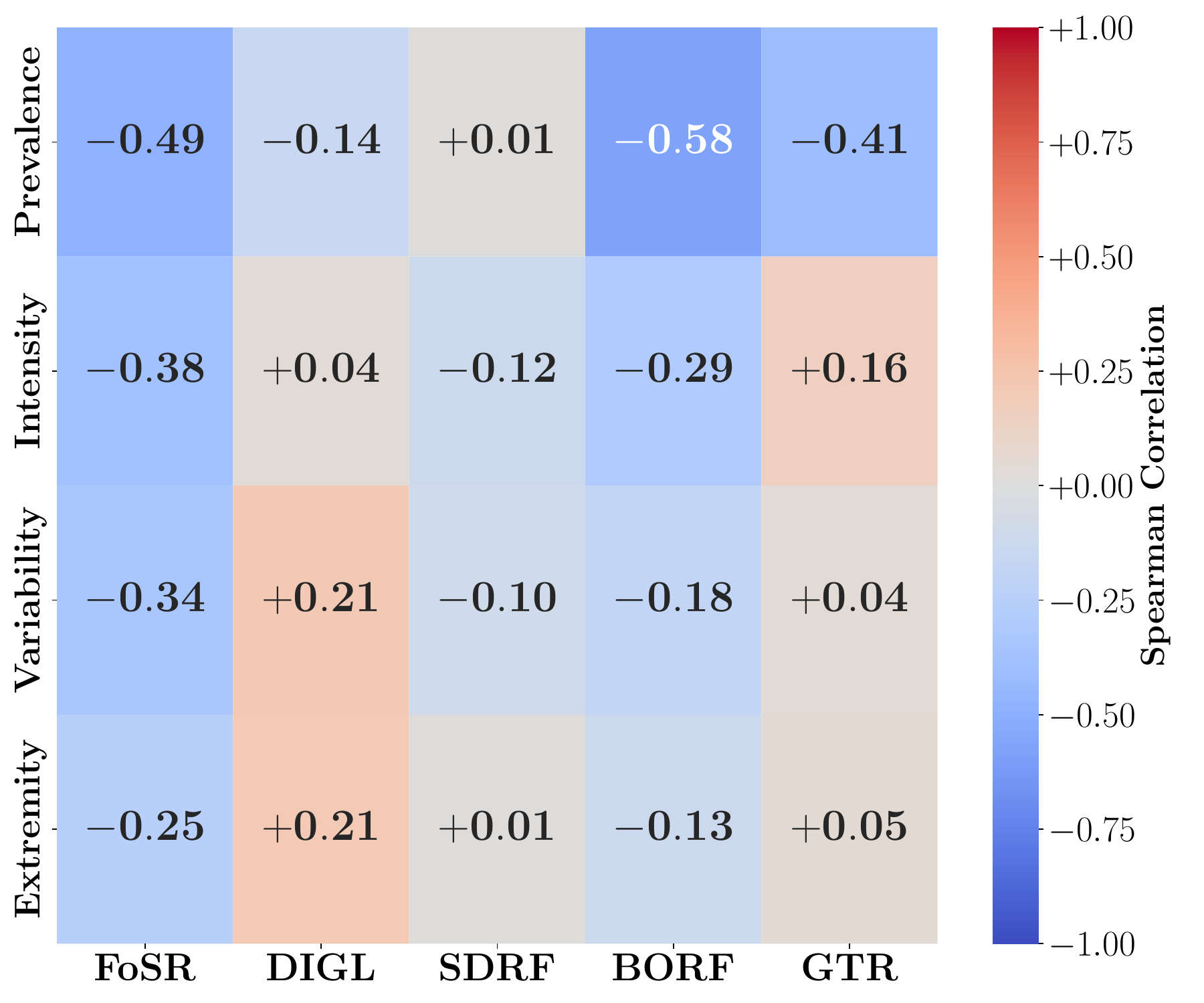}
        \caption{Graph classification tasks}
        \label{fig:heatmapgraph}
    \end{subfigure}
    \hspace{0.03\textwidth}
    \begin{subfigure}[b]{0.4\textwidth}
        \centering
        \includegraphics[width=\textwidth]{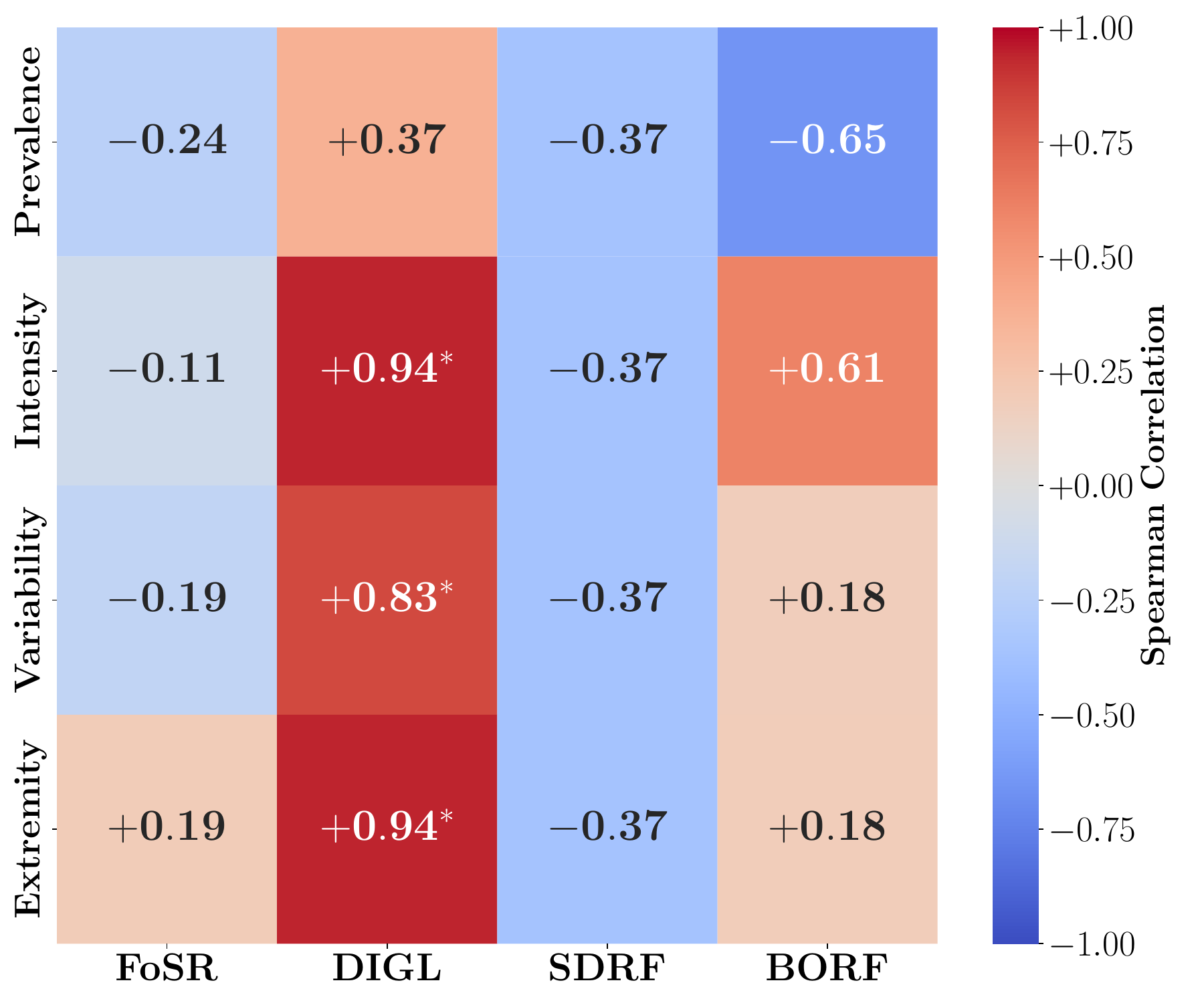}
        \caption{Node classification tasks}
        \label{fig:heatmapnode}
    \end{subfigure}
    \caption{Spearman correlation coefficients between treatment effects and performance gains for every metric–rewiring pair in (a) graph- and (b) node-classification tasks. Each coefficient is computed over all GNN-baseline hyperparameter configurations; an asterisk (*) indicates significance after multiple-comparison correction. Negative values imply that stronger mitigation (smaller treatment effects) aligns with larger performance gains---the desirable direction.}
    \label{fig:correlation_heatmaps}
\end{figure*}

% \vskip 0.03cm
% \noindent \textbf{Rewiring vs. Performance (Q3)}: To assess which rewiring strategy most effectively improves performance, we computed Spearman's correlation between each ATE metric and performance gain for every method, as shown in Figure~\ref{fig:heatmapgraph}.
% BORF emerged as the most effective, with reductions in prevalence, intensity, variability, and extremity of decay rates showing moderate correlations with performance gains---for instance, $-0.59$ for ATE on extremity and $-0.40$ for ATE on prevalence. 
% FoSR and SDRF exhibited similar trends: reductions in three out of four over-squashing metrics moderately correlated with performance improvement, peaking at $-0.39$ for ATE on prevalence (FoSR) and $-0.36$ for ATE on intensity (SDRF). In contrast, DIGL, despite achieving the largest reductions in over-squashing, did not yield performance gains---correlations were positive, suggesting that reducing ATE metrics was not aligned with accuracy improvement. This stems from DIGL's tendency to add a large number of edges, leading to over-smoothing, where node representations become indistinguishable. Prior studies have also documented DIGL’s over-smoothing issue by measuring the Dirichlet energy of node embeddings \cite{karhadkar2022fosr,choi2024panda}.

%%%% Amirali is here. 

\vskip 0.03cm
\noindent \textbf{Rewiring vs. Performance (Q3)}: To assess which rewiring strategy most effectively improves performance, we compute Spearman's correlation coefficient $\rho$ between each ATE metric and performance gain for every method, as shown in Figure~\ref{fig:heatmapgraph}. A negative correlation indicates a desirable outcome, where reductions in over-squashing (i.e., lower ATE values) are associated with improved generalization performance.
% Since lower ATE values indicate stronger reduction in over-squashing and higher performance gain is desirable, a negative correlation between ATE and performance gain reflects a beneficial relationship, where mitigating over-squashing translates into improved generalization.
FoSR emerged as the most effective in translating over-squashing mitigation to performance gains, showing mostly moderate correlations: prevalence ($\rho=-0.49$, moderate), intensity ($\rho=-0.38$, moderate) , variability ($\rho=-0.34$, moderate), and extremity ($\rho = -0.25$, weak).\footnote{The correlation strengths are based on Cohen's convention \cite{cohen2013statistical}: \emph{weak} for $\rho < 0.30$, \emph{moderate} for $0.30 \le \rho < 0.50$, and \emph{strong} for $\rho \ge 0.50$.}
BORF also shows negative (but comparatively weaker) correlations across all metrics: prevalence ($\rho = -0.58$, strong), intensity ($\rho = -0.29$, weak), variability ($\rho = -0.18$, weak), and extremity ($\rho = -0.13$, weak).
SDRF shows largely negligible and mixed correlations: prevalence ($\rho = +0.01$, weak/negligible), extremity ($\rho = +0.01$, weak/negligible), intensity ($\rho = -0.12$, weak), and variability ($\rho = -0.10$, weak).
GTR displays mostly weak positive correlations for intensity ($\rho = +0.16$, weak), variability ($\rho = -0.04$, weak/negligible), and extremity ($\rho = +0.05$, weak/negligible), but a negative correlation for prevalence ($\rho = -0.41$, moderate).
Although DIGL achieves the greatest over-squashing reduction (see Q2), three metrics show \emph{positive} correlations with performance gains, indicating that lower ATE values do not translate into higher performance. This paradox might be explained by  DIGL's heavy edge addition (see Table~\ref{tab:num_edges}), thus disrupting the graph’s original topology, weakening the local-message-passing inductive bias, and inducing over-smoothing---node representations converge and become indistinguishable.\footnote{Previous work has also linked DIGL to over-smoothing via elevated Dirichlet energy of node embeddings \cite{karhadkar2022fosr,choi2024panda}.}

Overall, only FoSR---and, to a lesser extent, BORF---translated reduced over-squashing into performance gains, whereas SDRF and GTR have negligible impact and DIGL’s notable reductions fail to improve performance, underscoring its susceptibility to over-smoothing and the disruption of the graph’s original topology and inductive bias.

\vskip 0.03cm
\noindent \textbf{Dataset Treatment Responsiveness (Q4)}:
Table~\ref{tab:responsiveness} presents the dataset-level responsiveness to rewiring---defined as the ratio of average treatment effects (across methods) to the original dataset over-squashing measurement. Negative values indicate over-squashing mitigation, where positive values show an increase in the over-squashing metric.
Social network datasets IMDB-B and Collab are the most responsive to rewiring. IMDB-B records the largest reductions in all metrics: prevalence by $-23.9\%$, intensity by $-38.5\%$, variability by $-70.1\%$, and extremity by $-25.5\%$. Collab follows closely, ranking second in prevalence ($-23.3\%$), intensity ($-28.1\%$), variability ($-28.4\%$), and extremity (-$25.3\%$). In contrast, Reddit-B resists mitigation the most: it ranks last across all metrics and is the only dataset where rewiring worsens over-squashing in intensity, variability, and extremity. We hypothesize that this is due to disconnected components within each graph, where rewiring inadvertently introduces new bottlenecks.

Among the bioinformatic datasets, Mutag exhibits the most consistent mitigation, with reductions in prevalence ($-20.2\%$), intensity ($-27.3\%$), variability ($-22.7\%$), and extremity ($-22.0\%$). Proteins is slightly less responsive, especially in prevalence ($-13.2\%$) and intensity ($-19.0\%$). Enzymes presents the weakest responsiveness among the three, with smaller and second-worst reductions in all four metrics: $-9.1\%$ in prevalence, $-16.9\%$ in intensity, $-20.2\%$ in variability, and $-18.5\%$ in extremity.

Overall, these results suggest that connected social graphs with dense community structure (Collab and IMDB-B) benefit most from rewiring, while large, disconnected networks such as Reddit-B---and to a lesser extent molecular graphs---pose greater challenges for over-squashing mitigation.\footnote{See the number of connected components of each dataset in Table \ref{tab:graphdatasetstats}, which supports this argument.} 
%%% These findings highlight the importance of dataset structure in determining the effectiveness of rewiring and underscore the need for diagnostic tools to assess over-squashing before applying mitigation. They also suggest that rewiring strategies should be adapted to dataset-specific properties to avoid counterproductive interventions.

\begin{table}[ht]
\caption{Responsiveness of each graph dataset to rewiring, reported in percentages.  Negative values denote desirable mitigation (metric decreases), whereas positive values indicate metric increases. The text color is \textcolor{green!75!black}{most}, \textcolor{blue}{second-most}, \textcolor{orange}{second-worst}, and \textcolor{red}{worst} responsive dataset.}
\label{tab:responsiveness}
\centering
\renewcommand{\arraystretch}{1.1}
\begin{tabularx}{1.0\columnwidth}{@{}l *{4}{>{\centering\arraybackslash}X}@{}}
\toprule
Dataset & Prevalence & Intensity & Variability & Extremity \\ \midrule
Mutag    & $-20.2$ & $-27.3$ & $-22.7$ & $-22.0$ \\
Proteins & $-13.2$ & $-19.0$ & $-27.6$ & $-21.0$ \\
Enzymes  & \textcolor{orange}{$-9.1$} & \textcolor{orange}{$-16.9$} & \textcolor{orange}{$-20.2$} & \textcolor{orange}{$-18.5$} \\
IMDB-B   & \textcolor{green!75!black}{$-23.9$} & \textcolor{green!75!black}{$-38.5$} & \textcolor{green!75!black}{$-70.1$} & \textcolor{green!75!black}{$-25.5$} \\
Collab   & \textcolor{blue}{$-23.3$} & \textcolor{blue}{$-28.1$} & \textcolor{blue}{$-28.4$} & \textcolor{blue}{$-25.3$} \\
Reddit-B & \textcolor{red}{$\! \! -4.0$}  & \textcolor{red}{$+96.9$}  & \textcolor{red}{$\ \ +101.1$} & \textcolor{red}{$\ \ +125.9$} \\
\bottomrule
\end{tabularx}
\end{table}

\begin{table*}[t]
\centering
\begin{threeparttable}
\caption{Treatment effects (ITEs) for node classification, averaged over each GNN baseline's hyperparameter configuration (GCN, GIN, R-GCN, and R-GIN). For every dataset–metric cell, background shading highlights the \colorbox{green!20}{best} and \colorbox{red!20}{worst} rewiring method. A green-filled circle (\gc) marks desirable negative ITEs, while a red-filled circle (\rc) marks undesirable positive ones. Gain is the percentage change in classification accuracy after rewiring. $\dagger$ marks not statistically significant results. Avg ITE reports the average treatment effect of each rewiring method across all datasets.}
\label{tab:noderesults}
\begin{tabularx}{\textwidth}{@{}l l *{4}{Y@{\hspace{-0.6em}}B}Y@{}}\toprule
\textbf{Rew.} & \textbf{Dataset} &
\multicolumn{8}{c}{\textbf{Individual Treatment Effect}} &
\textbf{Gain (\%)}\\
\cmidrule(lr){3-10}
& &
\textbf{Prevalence} & &
\textbf{Intensity}  & &
\textbf{Variability} & &
\textbf{Extremity} &
\\
\midrule
\multirow{7}{*}{FoSR}
 & Cora
    & $\phantom{-}6.0\mathrm{e}\text{-}4 \pm 2.2\mathrm{e}\text{-}2$ & \rc
    & \cellcolor{green!20}$\!-1.0\mathrm{e}\text{-}2 \pm 3.7\mathrm{e}\text{-}2$ & \gc
    & \cellcolor{green!20}$-4.5\mathrm{e}\text{-}3 \pm 2.8\mathrm{e}\text{-}2$ & \gc
    & \cellcolor{green!20}$\!-6.1\mathrm{e}\text{-}2 \pm 1.9\mathrm{e}\text{-}1$ & \gc
    & $-0.9 \pm 0.1$\\
 & Citeseer
    & \cellcolor{green!20}$-8.1\mathrm{e}\text{-}5 \pm 3.0\mathrm{e}\text{-}5$ & \gc
    & $\phantom{-}9.4\mathrm{e}\text{-}4 \pm 4.9\mathrm{e}\text{-}4$ & \rc
    & $\phantom{-}3.2\mathrm{e}\text{-}3 \pm 1.2\mathrm{e}\text{-}3$ & \rc
    & $\phantom{-}2.0\mathrm{e}\text{-}2 \pm 4.5\mathrm{e}\text{-}3$ & \rc
    & $\phantom{-}1.2 \pm 1.7$\\
 & Texas
    & $\phantom{-}7.5\mathrm{e}\text{-}2 \pm 5.0\mathrm{e}\text{-}2$ & \rc
    & $\phantom{-}1.6\mathrm{e}\text{-}2 \pm 1.3\mathrm{e}\text{-}3$ & \rc
    & $\phantom{-}1.2\mathrm{e}\text{-}2 \pm 9.9\mathrm{e}\text{-}4$ & \rc
    & $\phantom{-}9.1\mathrm{e}\text{-}2 \pm 5.1\mathrm{e}\text{-}2$ & \rc
    & $-2.3 \pm 5.9$\\
 & Cornell
    & \cellcolor{red!20}$\ \, \phantom{-}4.5\mathrm{e}\text{-}2 \pm 6.3\mathrm{e}\text{-}2^\dagger$ & \rc
    & \cellcolor{green!20}$\phantom{-}2.3\mathrm{e}\text{-}2 \pm 4.4\mathrm{e}\text{-}3$ & \rc
    & \cellcolor{green!20}$\phantom{-}1.9\mathrm{e}\text{-}2 \pm 9.7\mathrm{e}\text{-}2$ & \rc
    & \cellcolor{green!20}$\phantom{-}1.4\mathrm{e}\text{-}1 \pm 9.4\mathrm{e}\text{-}2$ & \rc
    & $-1.1 \pm 0.3$\\
 & Wiscon.
    & \cellcolor{red!20}$\phantom{-}4.4\mathrm{e}\text{-}2 \pm 5.1\mathrm{e}\text{-}2$ & \rc
    & \cellcolor{red!20}$\phantom{-}1.3\mathrm{e}\text{-}2 \pm 1.9\mathrm{e}\text{-}2$ & \rc
    & \cellcolor{red!20}$\phantom{-}1.1\mathrm{e}\text{-}2 \pm 1.7\mathrm{e}\text{-}2$ & \rc
    & \cellcolor{red!20}$\phantom{-}1.4\mathrm{e}\text{-}1 \pm 1.7\mathrm{e}\text{-}1$ & \rc
    & $\phantom{-}1.9 \pm 2.6$\\
 & Chamel.
    & $\phantom{-}1.8\mathrm{e}\text{-}1 \pm 1.0\mathrm{e}\text{-}2$ & \rc
    & $\phantom{-}8.5\mathrm{e}\text{-}3 \pm 3.8\mathrm{e}\text{-}3$ & \rc
    & $\phantom{-}2.8\mathrm{e}\text{-}2 \pm 3.5\mathrm{e}\text{-}4$ & \rc
    & $\phantom{-}1.1\mathrm{e}\text{-}1 \pm 1.6\mathrm{e}\text{-}2$ & \rc
    & $-0.9 \pm 1.2$\\
\cmidrule{2-11}
 & Avg ITE
    & $\phantom{-}5.7\mathrm{e}\text{-}2 \pm 6.6\mathrm{e}\text{-}2$ & \rc
    & $\phantom{-}8.6\mathrm{e}\text{-}3 \pm 1.1\mathrm{e}\text{-}2$ & \rc
    & $\phantom{-}1.1\mathrm{e}\text{-}2 \pm 1.1\mathrm{e}\text{-}2$ & \rc
    & $\phantom{-}7.3\mathrm{e}\text{-}2 \pm 7.9\mathrm{e}\text{-}2$ & \rc
    & $-0.4 \pm 1.4$\\
\midrule
\multirow{7}{*}{DIGL}
 & Cora
    & \cellcolor{red!20}$\! \! \phantom{-}6.4\mathrm{e}\text{-}1 \pm 0.0\mathrm{e}0$ & \rc
    & \cellcolor{red!20}$\! \! \phantom{-}2.0\mathrm{e}\text{-}1 \pm 0.0\mathrm{e}0$ & \rc
    & \cellcolor{red!20}$\! \! \phantom{-}1.6\mathrm{e}\text{-}1 \pm 0.0\mathrm{e}0$ & \rc
    & \cellcolor{red!20}$\phantom{-}2.2\mathrm{e}0 \pm 0.0\mathrm{e}0$ & \rc
    & $\phantom{-}1.3 \pm 0.0$\\
 & Citeseer
    & \cellcolor{red!20}$\! \! \phantom{-}1.8\mathrm{e}\text{-}1 \pm 0.0\mathrm{e}0$ & \rc
    & \cellcolor{red!20}$\! \! \phantom{-}1.4\mathrm{e}\text{-}1 \pm 0.0\mathrm{e}0$ & \rc
    & \cellcolor{red!20}$\! \! \phantom{-}1.5\mathrm{e}\text{-}2 \pm 0.0\mathrm{e}0$ & \rc
    & \cellcolor{red!20}$ \phantom{-}1.3\mathrm{e}0 \pm 0.0\mathrm{e}0$ & \rc
    & $\phantom{-}1.0 \pm 0.0$\\
 & Texas
    & \cellcolor{green!20}$\! \! \phantom{-}3.3\mathrm{e}\text{-}2 \pm 0.0\mathrm{e}0$ & \rc
    & \cellcolor{red!20}$\! \! \phantom{-}2.4\mathrm{e}\text{-}2 \pm 0.0\mathrm{e}0$ & \rc
    & \cellcolor{red!20}$\! \! \phantom{-}3.8\mathrm{e}\text{-}2 \pm 0.0\mathrm{e}0$ & \rc
    & \cellcolor{red!20}$\! \! \phantom{-}3.0\mathrm{e}\text{-}1 \pm 0.0\mathrm{e}0$ & \rc
    & $-0.8 \pm 0.0$\\
 & Cornell
    & $\! \! -3.0\mathrm{e}\text{-}2 \pm 0.0\mathrm{e}0$ & \gc
    & \cellcolor{red!20}$\! \! \phantom{-}2.0\mathrm{e}\text{-}1 \pm 0.0\mathrm{e}0$ & \rc
    & $\! \! \phantom{-}2.1\mathrm{e}\text{-}1 \pm 0.0\mathrm{e}0$ & \rc
    & $ \phantom{-}1.7\mathrm{e}0 \pm 0.0\mathrm{e}0$ & \rc
    & $\phantom{-}5.0 \pm 0.0$\\
 & Wiscon.
    & \cellcolor{green!20}$\! \! -4.0\mathrm{e}\text{-}1 \pm 0.0\mathrm{e}0$ & \gc
    & \cellcolor{green!20}$\! \! -7.4\mathrm{e}\text{-}3 \pm 0.0\mathrm{e}0$ & \gc
    & \cellcolor{green!20}$\! \! -7.4\mathrm{e}\text{-}3 \pm 0.0\mathrm{e}0$ & \gc
    & \cellcolor{green!20}$\! \! -1.0\mathrm{e}\text{-}1 \pm 0.0\mathrm{e}0$ & \gc
    & $-2.4 \pm 0.0$\\
 & Chamel.
    & \cellcolor{red!20}$\! \! \phantom{-}4.6\mathrm{e}\text{-}1 \pm 0.0\mathrm{e}0$ & \rc
    & \cellcolor{red!20}$\! \! \phantom{-}5.2\mathrm{e}\text{-}2 \pm 0.0\mathrm{e}0$ & \rc
    & \cellcolor{red!20}$\! \! \phantom{-}1.1\mathrm{e}\text{-}1 \pm 0.0\mathrm{e}0$ & \rc
    & \cellcolor{red!20}$\phantom{-}1.1\mathrm{e}0 \pm 0.0\mathrm{e}0$ & \rc
    & $-0.7 \pm 0.0$\\
\cmidrule{2-11}
 & Avg ITE
    & \cellcolor{red!20}$ \phantom{-}1.5\mathrm{e}\text{-}1 \pm 3.7\mathrm{e}\text{-}1$ & \rc
    & \cellcolor{red!20}$ \phantom{-}1.0\mathrm{e}\text{-}1 \pm 9.1\mathrm{e}\text{-}2$ & \rc
    & $\phantom{-}8.8\mathrm{e}\text{-}2 \pm 8.6\mathrm{e}\text{-}2$ & \rc
    & \cellcolor{red!20}$\ \phantom{-}1.1\mathrm{e}0 \pm 7.8\mathrm{e}\text{-}1$ & \rc
    & $\phantom{-}0.6 \pm 2.3$\\
\midrule
\multirow{7}{*}{SDRF}
 & Cora
    & \cellcolor{green!20}$\! \! \phantom{-}5.5\mathrm{e}\text{-}6 \pm 0.0\mathrm{e}0$ & \rc
    & $\! \! \phantom{-}5.3\mathrm{e}\text{-}5 \pm 0.0\mathrm{e}0$ & \rc
    & $\! \! \phantom{-}3.2\mathrm{e}\text{-}6 \pm 0.0\mathrm{e}0$ & \rc
    & $\! \! \phantom{-}1.9\mathrm{e}\text{-}5 \pm 0.0\mathrm{e}0$ & \rc
    & $\phantom{-}0.3 \pm 0.9$\\
 & Citeseer\tnote{*}
    &\ \phantom{-}N/A & &\ \phantom{-}N/A & &\ \phantom{-}N/A & &\ \phantom{-}N/A & &\ \phantom{-}\text{N/A}\\
 & Texas
    & \cellcolor{red!20}$\! \! \phantom{-}1.2\mathrm{e}\text{-}1 \pm 0.0\mathrm{e}0$ & \rc
    & $\! \! \phantom{-}2.2\mathrm{e}\text{-}3 \pm 0.0\mathrm{e}0$ & \rc
    & $\! \! \phantom{-}5.3\mathrm{e}\text{-}3 \pm 0.0\mathrm{e}0$ & \rc
    & $\! \! \phantom{-}8.7\mathrm{e}\text{-}2 \pm 0.0\mathrm{e}0$ & \rc
    & $-1.8 \pm 2.1$\\
 & Cornell\tnote{*}
    &\ \phantom{-}N/A & &\ \phantom{-}N/A & &\ \phantom{-}N/A & &\ \phantom{-}N/A & &\ \phantom{-}\text{N/A}\\
 & Wiscon.
    & $-5.2\mathrm{e}\text{-}2 \pm 1.2\mathrm{e}\text{-}1$ & \gc
    & $-1.2\mathrm{e}\text{-}3 \pm 2.9\mathrm{e}\text{-}3$ & \gc
    & $-5.1\mathrm{e}\text{-}4 \pm 2.2\mathrm{e}\text{-}3$ & \gc
    & $\phantom{-}6.8\mathrm{e}\text{-}3 \pm 4.0\mathrm{e}\text{-}2$ & \rc
    & $\phantom{-}0.4 \pm 0.4$\\
 & Chamel.
    & $-2.7\mathrm{e}\text{-}4 \pm 1.2\mathrm{e}\text{-}5$ & \gc
    & $-2.0\mathrm{e}\text{-}4 \pm 9.7\mathrm{e}\text{-}5$ & \gc
    & $-1.9\mathrm{e}\text{-}4 \pm 1.6\mathrm{e}\text{-}4$ & \gc
    & $\phantom{-}5.8\mathrm{e}\text{-}4 \pm 9.4\mathrm{e}\text{-}5$ & \rc
    & $\phantom{-}0.2 \pm 0.1$\\
\cmidrule{2-11}
 & Avg ITE
    & $\phantom{-}1.7\mathrm{e}\text{-}2 \pm 6.3\mathrm{e}\text{-}2$ & \rc
    & \cellcolor{green!20}$\phantom{-}2.1\mathrm{e}\text{-}4 \pm 1.2\mathrm{e}\text{-}3$ & \rc
    & \cellcolor{green!20}$\phantom{-}1.1\mathrm{e}\text{-}3 \pm 2.8\mathrm{e}\text{-}3$ & \rc
    & \cellcolor{green!20}$\phantom{-}2.4\mathrm{e}\text{-}2 \pm 3.7\mathrm{e}\text{-}2$ & \rc
    & $-0.2 \pm 0.9$\\
\midrule
\multirow{7}{*}{BORF}
 & Cora
    & $-6.8\mathrm{e}\text{-}5 \pm 1.4\mathrm{e}\text{-}6$ & \gc
    & $-8.0\mathrm{e}\text{-}4 \pm 1.0\mathrm{e}\text{-}4$ & \gc
    & $\phantom{-}1.7\mathrm{e}\text{-}4 \pm 2.7\mathrm{e}\text{-}5$ & \rc
    & $\phantom{-}1.2\mathrm{e}\text{-}2 \pm 2.1\mathrm{e}\text{-}4$ & \rc
    & $\phantom{-}2.3 \pm 2.1$\\
 & Citeseer
    & $\ \, \phantom{-}2.0\mathrm{e}\text{-}6 \pm 9.9\mathrm{e}\text{-}7$${}^{\dagger}$ & \rc
    & \cellcolor{green!20}$\ \, -3.3\mathrm{e}\text{-}6 \pm 2.3\mathrm{e}\text{-}6$${}^{\dagger}$ & \gc
    & \cellcolor{green!20}$-9.2\mathrm{e}\text{-}6 \pm 6.4\mathrm{e}\text{-}6$ & \gc
    & \cellcolor{green!20}$\phantom{-}0.0\mathrm{e}0 \pm 0.0\mathrm{e}0$ & \rc
    & $\phantom{-}2.7 \pm 1.6$\\
 & Texas
    & $\phantom{-}3.7\mathrm{e}\text{-}2 \pm 6.8\mathrm{e}\text{-}2$ & \rc
    & \cellcolor{green!20}$\phantom{-}7.6\mathrm{e}\text{-}4 \pm 8.7\mathrm{e}\text{-}4$ & \rc
    & \cellcolor{green!20}$\phantom{-}8.4\mathrm{e}\text{-}4 \pm 1.5\mathrm{e}\text{-}3$ & \rc
    & \cellcolor{green!20}$\phantom{-}2.6\mathrm{e}\text{-}2 \pm 4.7\mathrm{e}\text{-}2$ & \rc
    & $\phantom{-}7.4 \pm 3.1$\\
 & Cornell
    & \cellcolor{green!20}$-6.0\mathrm{e}\text{-}2 \pm 2.7\mathrm{e}\text{-}2$ & \gc
    & $\phantom{-}6.8\mathrm{e}\text{-}2 \pm 1.5\mathrm{e}\text{-}2$ & \rc
    & \cellcolor{red!20}$\phantom{-}6.9\mathrm{e}\text{-}1 \pm 3.9\mathrm{e}\text{-}1$ & \rc
    & \cellcolor{red!20}$\ \, \phantom{-}2.3\mathrm{e}0 \pm 7.8\mathrm{e}\text{-}2$ & \rc
    & $\, 10.7 \pm 2.0$\\
 & Wiscon.
    & $-3.1\mathrm{e}\text{-}2 \pm 1.1\mathrm{e}\text{-}2$ & \gc
    & $\phantom{-}3.3\mathrm{e}\text{-}4 \pm 1.0\mathrm{e}\text{-}4$ & \rc
    & $-9.5\mathrm{e}\text{-}4 \pm 5.4\mathrm{e}\text{-}4$ & \gc
    & $-9.9\mathrm{e}\text{-}3 \pm 2.9\mathrm{e}\text{-}3$ & \gc
    & $\phantom{-}6.1 \pm 0.5$\\
 & Chamel.
    & \cellcolor{green!20}$-8.2\mathrm{e}\text{-}3 \pm 1.4\mathrm{e}\text{-}5$ & \gc
    & \cellcolor{green!20}$-3.3\mathrm{e}\text{-}2 \pm 5.7\mathrm{e}\text{-}4$ & \gc
    & \cellcolor{green!20}$-2.1\mathrm{e}\text{-}2 \pm 2.8\mathrm{e}\text{-}4$ & \gc
    & \cellcolor{green!20}$-9.9\mathrm{e}\text{-}2 \pm 1.5\mathrm{e}\text{-}3$ & \gc
    & $\phantom{-}4.8 \pm 3.5$\\
\cmidrule{2-11}
 & Avg ITE
    & \cellcolor{green!20}$-1.1\mathrm{e}\text{-}2 \pm 3.4\mathrm{e}\text{-}2$ & \gc
    & $\phantom{-}5.8\mathrm{e}\text{-}3 \pm 3.0\mathrm{e}\text{-}2$ & \rc
    & \cellcolor{red!20}$\phantom{-}1.1\mathrm{e}\text{-}1 \pm 2.8\mathrm{e}\text{-}1$ & \rc
    & $\phantom{-}3.7\mathrm{e}\text{-}1 \pm 8.6\mathrm{e}\text{-}1$ & \rc
    & $\phantom{-}5.7 \pm 2.9$\\
\bottomrule
\end{tabularx}
\begin{tablenotes}[flushleft]
\footnotesize
\item[*] No edges are added by SDRF on Citeseer and Cornell; consequently, no treatment effect can be computed (entries marked ``N/A'').  
The performance change reported in the SDRF paper stems from a different hyperparameter set rather than the rewiring itself.
\end{tablenotes}
\end{threeparttable}
\end{table*}

\begin{table}[ht]
\caption{Responsiveness of each node dataset to rewiring, reported in percentages.  Negative values denote desirable mitigation (metric decreases), whereas positive values indicate metric increases. The text color is \textcolor{green!75!black}{most}, \textcolor{blue}{second-most}, \textcolor{orange}{second-worst}, and \textcolor{red}{worst} responsive dataset.}
\label{tab:responsiveness_node}
\centering
\renewcommand{\arraystretch}{1.1}
\begin{tabularx}{1.0\columnwidth}{@{}l *{4}{>{\centering\arraybackslash}X}@{}}
\toprule
Dataset & Prevalence & Intensity & Variability & Extremity \\
\midrule
Cornell   & \textcolor{blue}{$2.7$} 
          & \textcolor{orange}{$1079.0$} 
          & \textcolor{red}{$3865.3$} 
          & \textcolor{orange}{$1228.1$} \\
Texas     & $13.1$ 
          & $177.4$ 
          & $244.8$ 
          & $161.7$ \\
Wisconsin & \textcolor{green!75!black}{$\! \! \! \! -20.1$} 
          & \textcolor{blue}{$15.1$} 
          & \textcolor{green!75!black}{$6.6$} 
          & \textcolor{green!75!black}{$8.4$} \\
Cora      & \textcolor{orange}{$1052.6$} 
          & $129.5$ 
          & $150.6$ 
          & $264.7$ \\
Citeseer  & \textcolor{red}{$3260.9$} 
          & \textcolor{red}{$15210.4$} 
          & \textcolor{orange}{$346.6$} 
          & \textcolor{red}{$2391.3$} \\
Chameleon & $78.8$ 
          & \textcolor{green!75!black}{$4.8$} 
          & \textcolor{blue}{$34.6$} 
          & \textcolor{blue}{$71.1$} \\
\bottomrule
\end{tabularx}
\end{table}

\subsection{Results on Node Classification Tasks}
We report our results answering Q1--Q4 for node classification. 

\vskip 0.03cm
\noindent \textbf{Dataset Over-Squashing Levels (Q1)}: Table \ref{tab:nodedatasetstats} shows that over-squashing is generally weak across datasets and metrics. Cornell, Texas, and Wisconsin display the highest prevalence ($0.50\text{--}0.55$, large), but with low intensity ($0.006$–$0.009$), variability ($0.005$–$0.008$), and extremity ($0.08$–$0.11$), indicating widespread yet mild compression. Chameleon has the highest intensity ($0.14$, moderate) and extremity ($0.39$, strong), but with low prevalence ($20\%$, small), suggesting severe, uneven bottlenecks over a small subset of pairs---making it a suitable benchmark for mitigation studies. Cora and Citeseer show the lowest prevalence ($0.015$ and $0.002$, respectively), along with the lowest intensity and variability, indicating minimal over-squashing. A cross-examination of Tables \ref{tab:graphdatasetstats} and \ref{tab:nodedatasetstats} shows graph-task datasets are more susceptible to over-squashing compared to node-task datasets.

\vskip 0.05cm
\noindent \textbf{Rewiring Effectiveness (Q2)}:
Table~\ref{tab:noderesults} shows in node-classification tasks, rewiring more often \emph{increases} over-squashing (red markers, \rc) than reduces it (green markers, \gc)---the opposite of the trend observed in graph-classification benchmarks.
FoSR generally raises the metrics, with only a few exceptions---Cora shows a slight decrease (intensity -0.01, variability +0.004, extremity -0.06) and Citeseer a negligible drop in prevalence (-0.0001)---all of which are weak treatment effects. DIGL also increases over-squashing in five of the six datasets; Wisconsin is the exception with the weak treatment effects (intensity -0.007, variability -0.007, extremity -0.1).
SDRF has a near-zero impact, with changes mostly on the order of $10^{-6}$ to $10^{-2}$). These negligible effects have improved over-squashing just for Wisconsin and Chameleon. BORF exhibits mixed behavior, improving over-squashing for some metrics in some datasets while worsening others.  

Ranking the methods by treatment effects shows that DIGL performs the worst in most datasets and metrics---except in Wisconsin, where it surprisingly ranks best across all metrics. BORF ranks best for intensity, variability, and extremity in Citeseer, Texas, and Chameleon, while FoSR is best for these metrics in Cora and Cornell. In Wisconsin, however, FoSR is the worst overall metric. 

Aggregating effects across datasets for each method (AVG ITE in Table \ref{tab:noderesults}), DIGL exhibits the strongest adverse effects (prevalence +0.15, intensity +0.1, variability +0.09, and extremity +1.1 on average). In all metrics except variability, DIGL is worse than other methods. SDRF has a near-negligible impact: it slightly increases all four metrics (worsens over-squashing), yet its increments in intensity, variability, and extremity are the smallest among the other methods, making it the ``least harmful'' of the rewiring options. BORF shows a mixed pattern: it slightly lowers prevalence (-0.011, the best among all methods) but sharply increases extremity (+0.37) and variability (+0.11, the worst), indicating reduced global compression at the cost of new local bottlenecks.

Overall, as node-classification benchmarks are structurally less prone to over-squashing (Table \ref{tab:nodedatasetstats}), aggressive (e.g., DIGL) or even moderate (e.g, SDRF and BORF) rewiring is often ineffective or counterproductive. While added connectivity relieves bottlenecks in graph-classification benchmarks with high over-squashing, it often disrupts local structure in node-classification benchmarks with low over-squashing, creating new compression pathways.

\vskip 0.05cm
\noindent \textbf{Rewiring vs. Performance (Q3)}: 
Figure \ref{fig:heatmapnode} shows the correlation coefficient $\rho$ between each rewiring method's treatment effect and its performance changes. DIGL shows strong, significant positive correlations for three metrics ($\rho \geq 0.83$), suggesting its performance gains coincide with increased over-squashing. SDRF shows moderate, non-significant negative correlations ($\rho = -0.37$); while in the ``right'' direction, effects are too small to yield meaningful gains. FoSR exhibits weak, non-significant negative correlations for most metrics, indicating little potential performance gain. BORF mostly shows non-significant positive correlations, implying its mitigation may drop the performance. Overall, in node-classification benchmarks, rewiring rarely improves performance by reducing over-squashing. On the contrary, performance gains---particularly in DIGL---often coincide with increased compression, suggesting that other mechanisms (e.g., altered propagation patterns or smoothing behavior) drive the improvements.
% Figure \ref{fig:heatmapnode} shows the correlation coefficient $\rho$ between each rewiring method's treatment effect and the resulting performance changes. FoSR exhibits weak, non-significant negative correlations across all metrics (e.g., $\rho = -0.11$ for intensity), indicating little potential performance gain via over-squashing mitigation. DIGL, surprisingly, shows strong, significant positive correlations for intensity, variability, and extremity ($\rho = 0.94$, for intensity and extremity, and $\rho = 0.83$ for variability), meaning its performance gains actually coincide with increased over-squashing. SDRF displays non-significant, moderate negative correlations ($\rho = -0.37$); although the correlations are in the ``right'' directions, the treatment effects are so small that practical gains are negligible. BORF shows non-significant positive correlations with intensity, variability, and extremity, implying its mitigation tends to drop the performance. Overall, these findings reinforce that, for node-classification benchmarks, rewiring rarely improves performance by reducing over-squashing. On the contrary, performance gains---particularly in DIGL---often coincide with increased compression, suggesting that other mechanisms (e.g., altered propagation patterns or smoothing behavior) are responsible for the observed generalization benefits.

\vskip 0.03cm
\noindent \textbf{Dataset Treatment Responsiveness (Q4)}: Table~\ref{tab:responsiveness_node} presents the dataset-level responsiveness to over-squashing.  
% ---defined as the ratio of average treatment effects (across methods) to the original dataset over-squashing measurement.
In contrast to the graph classification setting, none of the node classification datasets exhibit ``true'' responsiveness to over-squashing mitigation. All values (except one) are positive, indicating that rewiring methods not only fail to reduce over-squashing but often exacerbate it for any dataset under any metric. Citeseer is the extreme case, with dramatic increases across all metrics---prevalence rises by $3260\%$, intensity by $15210\%$, variability by $346\%$, and extremity by $2391\%$.
% Wisconsin records a small reduction in prevalence ($-18.3\%$) and the lowest increase in extremity ($9.2\%$), though its intensity and variability still grow substantially. Cornell and Texas both experience consistent increases in all metrics: Cornell with a modest growth in prevalence ($1.8\%$), but triple-digit spikes in other metrics, and Texas with intensity growth of $178\%$ and a 17-fold spike in its variability.
% Chameleon, while slightly more stable, still shows positive values, with a relatively low intensity increase ($14.1\%$), but over $100\%$ growth in variability.
Cornell ranks as the second worst, with consistent increases across all metrics. Texas also performs poorly, showing a 177\% rise in intensity and a $244\%$ increase in variability. While Chameleon and Wisconsin appear more stable, both still show metric growth, suggesting they are poor candidates for over-squashing analysis.
These results suggest that rewiring often introduces new bottlenecks in node classification tasks, rather than relieving them. Since most of these datasets already operate in a regime with relatively weak over-squashing, the benefits of rewiring are limited and may even be counterproductive.

\subsection{Discussion: Graph vs. Node Classification}
\noindent \textbf{Dataset Over-Squashing Levels (Q1)}: A cross-task comparison shows that graph-classification datasets exhibit substantially higher levels of over-squashing than node-classification ones (compare Tables \ref{tab:graphdatasetstats} and \ref{tab:nodedatasetstats}).
Graph-level benchmarks show high prevalence in five out of six datasets, with the remaining one at a moderate level. The same pattern exists for extremity. For intensity and variability, only two datasets fall in the weak/low range, while the rest exhibit moderate or strong/high values. In contrast, node datasets are largely unaffected: apart from Chameleon's strong extremity and the high prevalence in Cornell, Texas, and Wisconsin, most metrics across all datasets remain weak, low, or small. Hence, over-squashing is a core obstacle in graph classification datasets, but a more localized or negligible issue in node datasets---implying that mitigation efforts may be more impactful and necessary for the former.

\vskip 0.03cm
\noindent \textbf{Rewiring Effectiveness (Q2)}:
Rewiring effectively mitigates over-squashing in graph classification benchmarks, but is often harmful in node classification datasets. In graph datasets, added connectivity---especially via DIGL’s dense rewiring---consistently reduces all over-squashing metrics. In node datasets, the same interventions frequently worsen over-squashing, as already balanced or mildly compressed structures (i.e., graphs with minimal over-squashing) are disrupted, and new bottlenecks emerge. This suggest that rewiring is beneficial only when over-squashing is severe or moderate (e.g., graph-classification datasets) and can be counter-productive when over-squashing is low (most node-classification graphs).

\vskip 0.03cm
\noindent \textbf{Rewiring vs. Performance (Q3)}:
% A cross-task comparison shows that the relationship between over-squashing reduction and performance improvement differs in graph and node classification datasets.
In graph-level benchmarks, methods such as FoSR and BORF show negative correlations between reduced over-squashing and improved accuracy, confirming that alleviating information bottlenecks enhances generalization. DIGL, despite achieving strong over-squashing reductions, fails to improve performance---likely because its aggressive edge additions erase topological information.
In node-level tasks, DIGL and BORF often show performance gains while surprisingly increasing over-squashing, suggesting that other factors (e.g., smoothing or altered message propagation) drive the gains. SDRF and FoSR exhibit no correlations.
These findings suggest that rewiring mitigation helps performance gain when over-squashing is pronounced (as in most graph classification datasets) and not overcorrecting (as in FoSR or BORF). By contrast, when over-squashing is mild---as in typical node-classification graphs---rewiring rarely converts metric improvements into accuracy gains.

\vskip 0.03cm
\noindent \textbf{Dataset Treatment Responsiveness (Q4)}:
Dataset-level responsiveness to rewiring is pronounced in graph classification datasets but not in node classification tasks. 
In graph classification datasets, well-connected social networks (i.e., Collab and IMDB-B) exhibit the strongest responsiveness, bioinformatic graphs are moderately responsive, and Reddit-B, with the most disconnected components, is the least responsive dataset. In node datasets, rewiring rarely helps and often hurts. Wisconsin is the only dataset with meaningful responsiveness in prevalence. The rest show counter-responsiveness: Citeseer and Cornell most strongly, followed by Cora and Texas, with Chameleon and Wisconsin to a lesser degree. These findings suggest rewiring pays off only when over-squashing is severe and global, but offers limited or negative impact when compression is mild, underscoring the need for dataset-aware interventions.

\section{Conclusion and Future Work}
We proposed topology-focused measurements for over-squashing, built on its formal characterization by modeling the exponential decay of node-pair sensitivity with increasing network depth. %This framework enables an empirical, topology-based measure of pairwise over-squashing.
We extend our measurements to the graph-level measures and integrate them into a causal inference framework to evaluate the effect of rewiring on over-squashing. We conduct extensive empirical analyses of over-squashing and its rewiring-based mitigation strategies on a wide range of real-world benchmarks for both node- and graph-classification tasks. 
Our analyses addressed four research questions: (i) the extent of over-squashing, (ii) how effectively rewiring mitigates it, (iii) whether such mitigation leads to performance gains, and (iv) how responsive each dataset is to over-squashing reduction. We found that graph-classification datasets (except Reddit-B) suffer substantial over-squashing and are generally responsive: rewiring lowers our metrics and often boosts accuracy. Node-classification benchmarks show little over-squashing; rewiring often \emph{increases} compression, and its performance effects are largely unrelated to over-squashing. Our findings underscore the importance of applying rewiring selectively, based on the presence of over-squashing. Future work includes extending our experiments to dynamic rewiring methods, exploring the relationship between negative decay rates and over-smoothing, and designing novel rewiring methods guided by our over-squashing measurement framework.

\section*{GenAI Usage Disclosure}
We used Generative AI tools in a limited capacity to support this work. During the writing process, GenAI was employed occasionally to improve the clarity and readability of the text, such as for grammar checking and suggesting alternative word choices. In terms of coding, GenAI tools were used primarily to assist with writing documentation and resolving debugging issues. All core research contributions, including the development of the methodology, design of experiments, analysis of results, and interpretation of findings, were conceived and executed independently by the authors without the involvement of GenAI systems.
\bibliographystyle{ACM-Reference-Format}
\bibliography{main}

\appendix
% \section*{Appendix}
% \vspace{0.4cm}
\section{Normalized Jacobian Norm Approximation}
\label{app:relativeapprox}
\begin{proof}
We derive an approximation of the relative Jacobian norm \(\mathcal{\tilde{J}}_\ell(v, u)\) under the assumption of a linear message-passing GNN. Let the $\ell$-th layer of a \emph{linear} message-passing GNN be
\begin{equation}
    \mathbf{H}^{(\ell)} = \mathbf{\tilde{A}}^{}\mathbf{H}^{(\ell-1)}\mathbf{W}^{\ell}
\end{equation}
where $\mathbf{\tilde{A}} = \mathbf{A} + \mathbf{I}$ is the self-loop–augmented adjacency matrix, $\mathbf{H}^{(\ell-1)}$ stacks node embeddings of the layer $\ell-1$ as rows, and $\mathbf{W}^{\ell}$ is the learnable weight matrix of the $\ell$-th layer. Iterating from the initial features $\mathbf{H}^{0}$ gives
\begin{equation}
    \mathbf{H}^{(\ell)} = \mathbf{\tilde{A}}^\ell \mathbf{H}^{(0)}\mathbf{W}, \qquad \mathbf{W}:= \mathbf{W}^{(1)}\mathbf{W}^{(2)}\dots \mathbf{W}^{(\ell)}.
\end{equation}
For any node $v$, its representation after $\ell$ layers is:
\begin{equation}
    \mathbf{h}_v^{(\ell)} = \sum^n_{u=1}(\mathbf{\tilde{A}^\ell})_{uv}  \mathbf{h}_u^{(0)}\mathbf{W},
\end{equation}
where $\mathbf{h}_u^{(0)}$  is the input feature row of node $u$. The Jacobian of \(\mathbf{h}_v^{(\ell)}\) with respect to \(\mathbf{h}_u^{(0)}\) is then
\begin{equation}
    \frac{\partial\mathbf{h}_v^{(\ell)}}{\partial \mathbf{h}_u^{(0)}} = (\mathbf{\tilde{A}^{\ell}})_{uv}\mathbf{W} \in \mathbb{R}^{d_0 \times d_{\ell}},
\end{equation}
Given that the matrix norms are homogeneous---for any scaler $c$, $\|c\ \mathbf{W}\|$ = $|c| \|\mathbf{W}\|$---we compute the Frobenius norm
\begin{equation}
    \left\|\frac{\partial\mathbf{h}_v^{(\ell)}}{\partial \mathbf{h}_u^{(0)}}\right\| = \|(\mathbf{\tilde{A}}^\ell)_{uv}\mathbf{W}\| = (\mathbf{\tilde{A}}^\ell)_{uv} \|\mathbf{W}\|.
\end{equation}

To compute the relative Jacobian norm, we normalize by the 
total sensitivity of $v$ to all input nodes:
\begin{equation}
    \sum_k\left\|\frac{\partial\mathbf{h}_v^{(\ell)}}{\partial \mathbf{h}_k^{(0)}}\right\|  = \|\mathbf{W}\| \sum_{k}(\mathbf{\tilde{A}}^\ell)_{kv}.
\end{equation}
Canceling the common factor $\|\mathbf{W}\|$, the relative Jacobian norm simplifies to the exact expression:
\begin{equation}
    \mathcal{\tilde{J}}_\ell(v, u) = \frac{(\mathbf{\tilde{A}}^\ell)_{uv}}{\sum_k (\mathbf{\tilde{A}}^\ell)_{kv}}.
\end{equation}
\end{proof}
%Amirali is here.
\begin{figure*}[h!]
    \centering
    \begin{subfigure}[b]{0.246\textwidth}
        \centering
        \includegraphics[width=\textwidth]{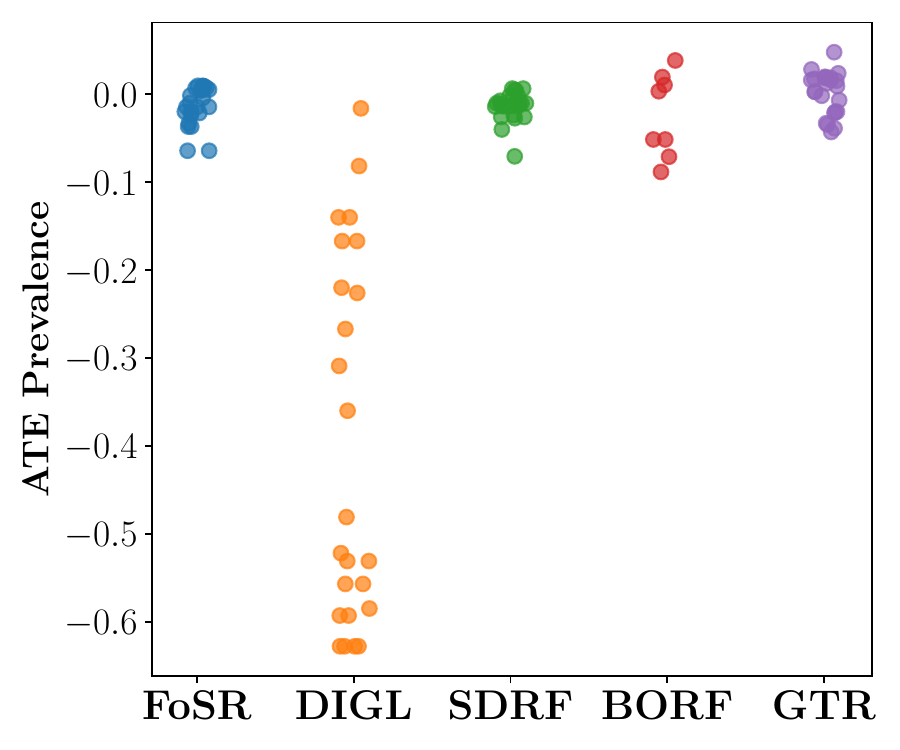}
        \caption{Prevalence}
    \end{subfigure}
    %\hspace{0.03\textwidth}
    \begin{subfigure}[b]{0.246\textwidth}
        \centering
        \includegraphics[width=\textwidth]{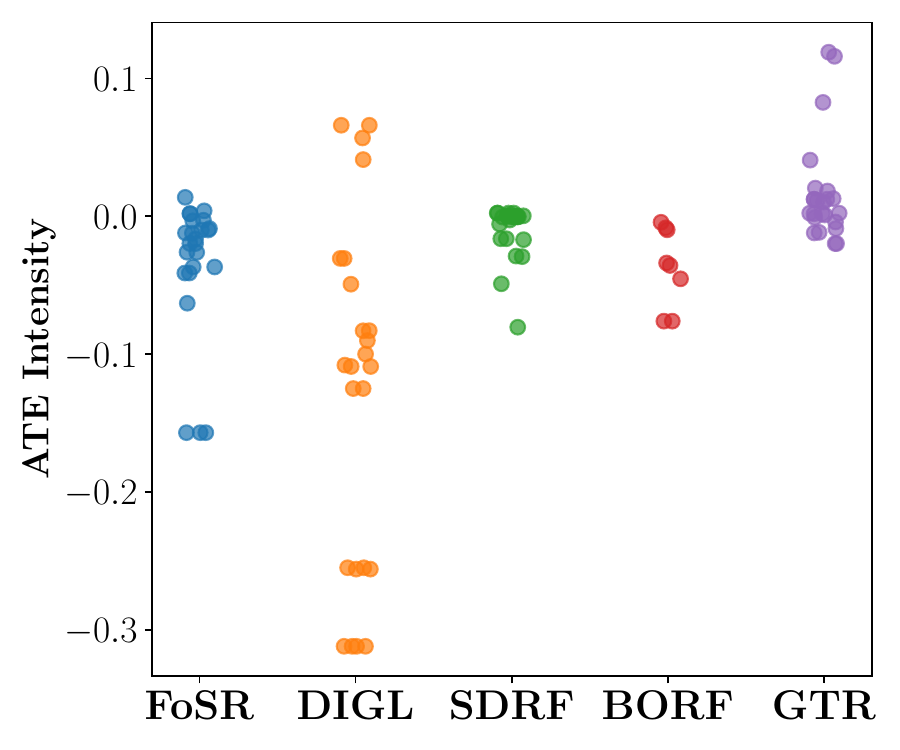}
        \caption{Intensity}
    \end{subfigure}
    %\vspace{0.8\baselineskip}
    \begin{subfigure}[b]{0.246\textwidth}
        \centering
        \includegraphics[width=\textwidth]{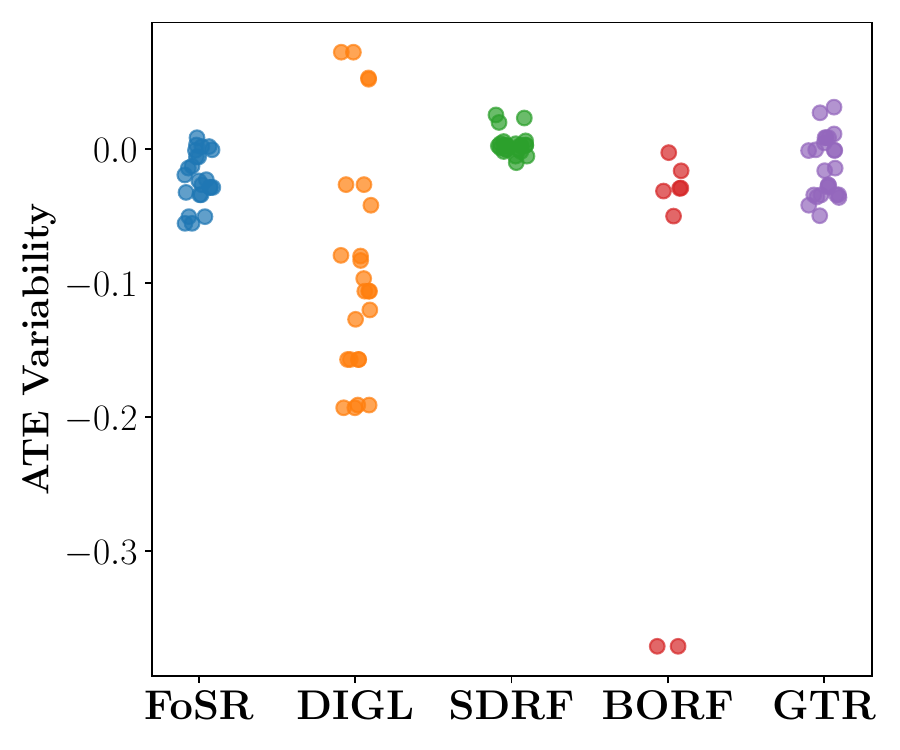}
        \caption{Variability}
    \end{subfigure}
    %\hspace{0.03\textwidth}
    \begin{subfigure}[b]{0.246\textwidth}
        \centering
        \includegraphics[width=\textwidth]{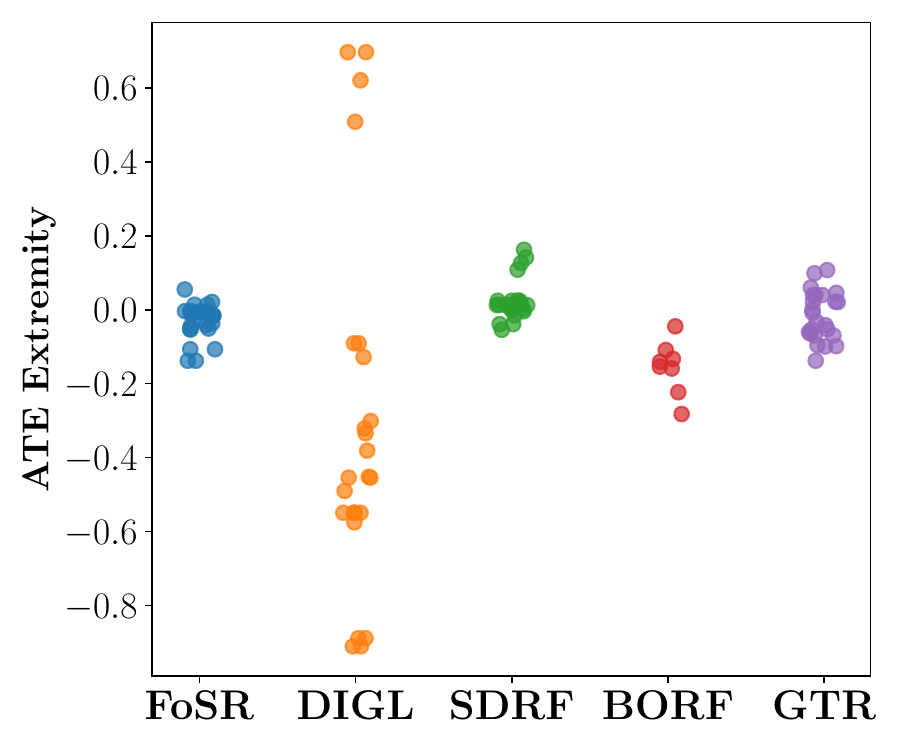}
        \caption{Extremity}
    \end{subfigure}
    \caption{Scatter plots of $\text{ATEs}$ for each rewiring strategy on graph-classification tasks under metrics of prevalence, intensity, variability, and extremity. Each point represents the result from a specific GNN-baseline hyperparameter configuration.}
    \label{fig:graphscatter}
\end{figure*}

\begin{figure*}[h!]
    \centering
    \begin{subfigure}[b]{0.246\textwidth}
        \centering
        \includegraphics[width=\textwidth]{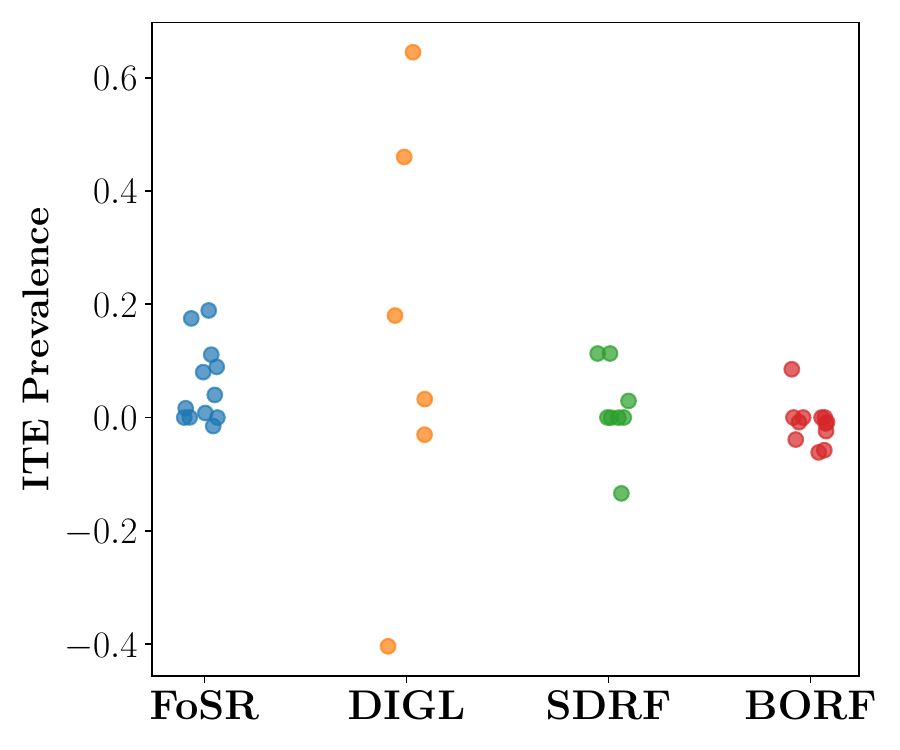}
        \caption{Prevalence}
    \end{subfigure}
    %\hspace{0.03\textwidth}
    \begin{subfigure}[b]{0.246\textwidth}
        \centering
        \includegraphics[width=\textwidth]{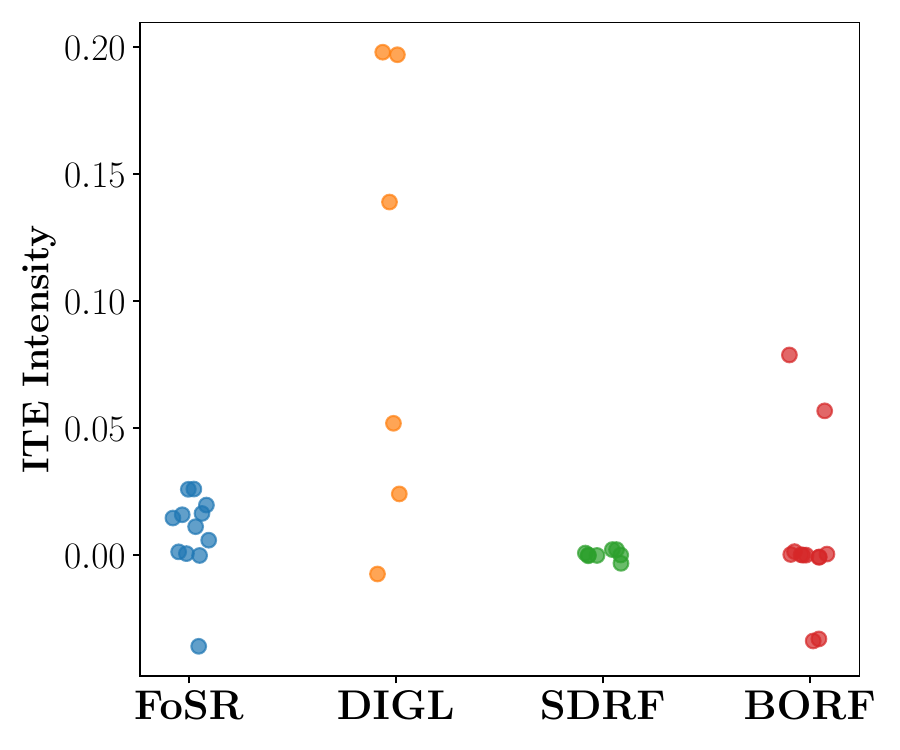}
        \caption{Intensity}
    \end{subfigure}
    %\vspace{0.8\baselineskip}
    \begin{subfigure}[b]{0.246\textwidth}
        \centering
        \includegraphics[width=\textwidth]{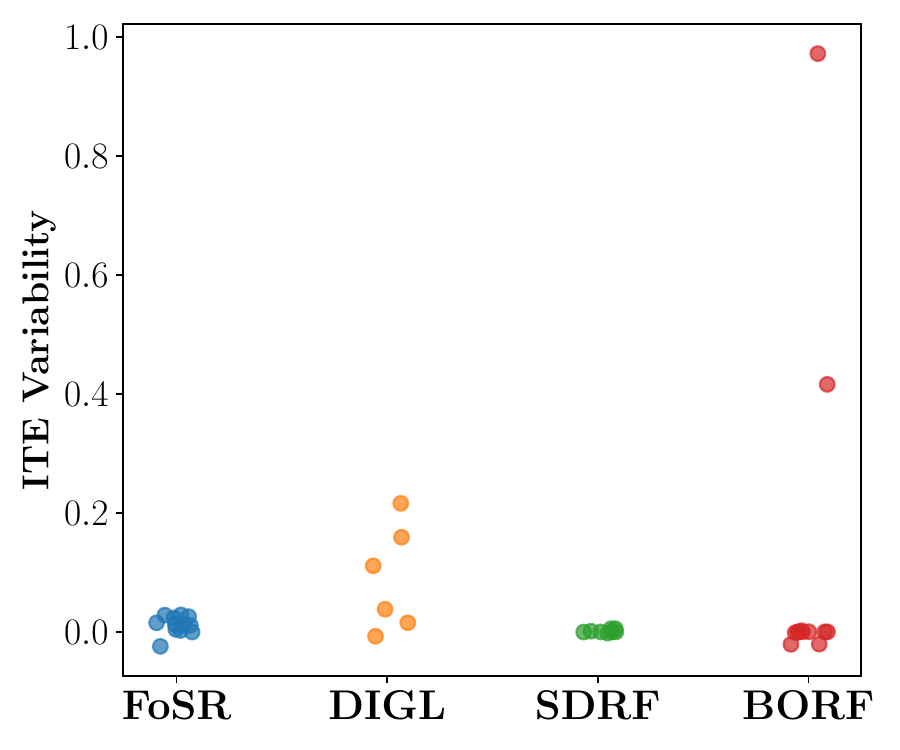}
        \caption{Variability}
    \end{subfigure}
    %\hspace{0.03\textwidth}
    \begin{subfigure}[b]{0.246\textwidth}
        \centering
        \includegraphics[width=\textwidth]{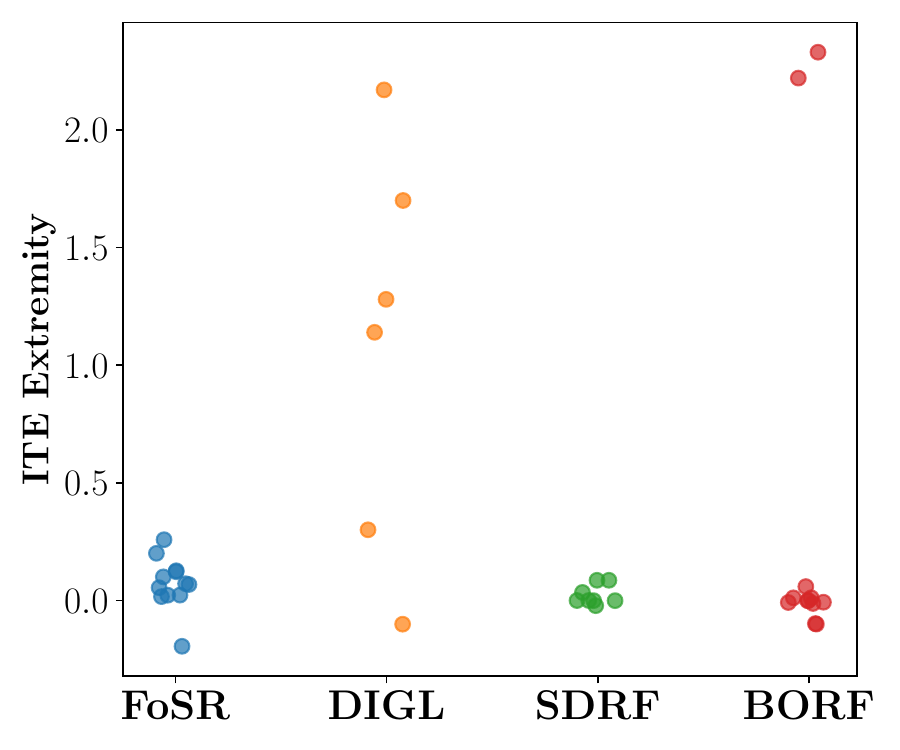}
        \caption{Extremity}
    \end{subfigure}
    \caption{Scatter plots of $\text{ITEs}$ for each rewiring strategy on node-classification tasks under metrics of prevalence, intensity, variability, and extremity. Each point represents the result from a specific GNN-baseline hyperparameter configuration.}
    \label{fig:scatternode}
\end{figure*}

\section{Causal Assumptions and Framework Validity}
\label{app:causality}
In causal inference, several key assumptions must be held to draw valid conclusions. We, therefore, examine each assumption to confirm our framework can reliably estimate the causal effect of rewiring on over-squashing.

\vskip 0.5mm
\noindent \emph{The Stable Unit Treatment Value Assumption (SUTVA)} requires that the treatment assigned to one unit does not affect the outcome of another. In our setup, each graph is an independent unit and is treated independently, and rewiring one graph cannot influence another, so no interference occurs---satisfying SUTVA.

\vskip 0.5mm
\noindent\emph{Positivity} requires that each graph has a non-zero probability of receiving both treatment conditions: rewiring and no rewiring.
This entails (i) feasibility of treatment---every graph must be rewirable using the chosen method, and (ii) treatment assignment variation---a sufficient number of graphs should be in both the treated and untreated groups. These conditions are met in our setting, where many graphs remain unchanged by rewiring (effectively untreated), ensuring overlap.

\vskip 0.5mm
\noindent\emph{Exchangeability} (also known as ignorability/unconfoundedness) requires that no unmeasured confounders affect the treatment and the outcome. Our within-unit design compares the same graph before and after rewiring, holding all fixed graph properties constant. This implicitly controls for confounders tied to the graph’s structure, as any change in over-squashing must stem from the rewiring intervention itself. This means that any differences in the outcome can reasonably be attributed to the rewiring, not to differences in the graph’s inherent traits---satisfying exchangeability for time-invariant confounders.

\vskip 0.5mm
\noindent\emph{Consistency} requires that the treatment is applied consistently across all units, and the over-squashing metrics to be computed identically pre- and post-treatment. This ensures the observed outcomes correspond to the intended treatment effect---a condition satisfied in our framework.

\vskip 0.5mm
\noindent We additionally assume \emph{No Systematic Measurement Error} in over-squashing metrics and \emph{Independence of Observations}, as graphs are distinct units. Lastly, our sample size is sufficient to ensure statistical power for detecting meaningful causal effects.

These assumptions support the validity of our causal inference framework, enabling a rigorous assessment of rewiring’s impact on over-squashing.

\section{Scatter Plot View of Treatment Effects}
Figures~\ref{fig:graphscatter} and \ref{fig:scatternode} illustrate the ATEs and ITEs (respectively) of each rewiring method on the graph- and node-classification (respectively) datasets. As also shown in Table~\ref{tab:graphresults}, DIGL most effectively mitigates over-squashing across graph classification tasks, outperforming the other rewiring strategies. In contrast, the reductions achieved by FoSR, SDRF, GTR, and BORF are more moderate. For node-classification datasets, however, the changes are predominantly positive---indicating increased over-squashing. DIGL performs the worst in this setting, amplifying over-squashing metrics more than the other methods. The remaining three baselines show more comparable performance, though BORF tends to produce more outliers than FoSR and SDRF. These findings highlight that rewiring is effective in reducing over-squashing when over-squashing is pronounced in the dataset---as in most graph-classification datasets---whereas, in datasets with small over-squashing, the same interventions may inadvertently amplify it.

\section{Number of Added Edges After Rewiring}
Table \ref{tab:num_edges} reports the average number of edges added by each rewiring method for each dataset. In graph classification datasets, which contain multiple graphs, we (i) compute the number of edges added in each graph, (ii) average these values across all graphs in the dataset, and (iii) average again across every hyperparameter configuration under which the rewiring method has been evaluated in prior studies. In node classification datasets, each dataset consists of a single graph, so we only perform steps (i) and (iii). 
Table \ref{tab:num_edges} shows three distinct rewiring behaviors, differing in how aggressively each method modifies the graph across both graph- and node-classification tasks..
DIGL is the most aggressive method. In graph-classification datasets, it adds a large number of edges---ranging from $138$ on Mutag to $47,000$ on Reddit-B---often exceeding others by orders of magnitude. This pattern is even more pronounced in node datasets: DIGL adds over $150k$ and $100k$ (respectively) edges to Cora and  Citeseer (respectively). While this level of augmentation drastically increases graph connectivity, it also raises the risk of over-smoothing, scalability, and loss of topology-defined inductive bias. FoSR and GTR follow a moderate and fixed-budget strategy. In graph classification datasets, they add a consistent number of $8-33$ edges per graph on average---regardless of dataset size. On node classification datasets, FoSR adds $100$ or $150$ edges, reflecting its deterministic, size-invariant design. SDRF and BORF apply minimal rewiring. On graph datasets, SDRF consistently adds fewer than $5$ edges per graph, while BORF varies more: it may add a moderate number (e.g., $25.7$ on Enzymes) or remove more edges than it adds, as seen with a net change of $–58.7$ on IMDB-B. On node classification datasets, both methods exhibit similarly restrained behavior: SDRF adds little or nothing in many cases, and BORF frequently results in zero or negative changes (e.g., $-12.5$ on Cornell, $-10.0$ on Chameleon). These patterns suggest that both methods prioritize minimal intervention and, in BORF’s case, actively favor edge pruning when beneficial.
This analysis highlights three distinct rewiring philosophies: DIGL prioritizes connectivity maximization, FoSR and GTR add edges conservatively but consistently, and SDRF/BORF adopt a highly selective, sparsity-preserving approach.

\begin{table*}[t]
\caption{For graph classification datasets, values represent the average number of added edges per graph, computed across all graphs in the dataset and averaged over the GNN baseline hyperparameter configurations for each rewiring technique. For node classification datasets, the values are averaged over all hyperparameter configurations of the specific GNN baseline used for evaluating the rewiring technique.}
\label{tab:num_edges}
\centering
\begin{threeparttable}
\setlength{\tabcolsep}{1.8mm}
\renewcommand{\arraystretch}{1.1}
\begin{tabularx}{\textwidth}{@{} l *{6}{>{\centering\arraybackslash}X} @{}}
\toprule
\textbf{Rewiring} & \multicolumn{6}{c}{\textbf{Graph Classification Datasets}} \\
\cmidrule(lr){2-7}
 & \textbf{Mutag} & \textbf{Proteins} & \textbf{Enzymes} & \textbf{IMDB-B} & \textbf{Reddit-B} & \textbf{Collab} \\
\midrule
FoSR & $26.2$ & $13.7$ & $23.7$ & $16.2$ & $15.0$ & $11.2$ \\
DIGL & $138.3$ & $740.5$ & $390.6$ & $139.6$ & $47530.6$ & $2102.4$ \\
SDRF & $0.8$ & $3.3$ & $2.1$ & $2.1$ & $5.5$ & $17.5$ \\
GTR & $33.8$ & $11.3$ & $28.8$ & $20.0$ & $21.3$ & $8.8$ \\
BORF\tnote{*} & $3.5$ & $12.36$ & $25.7$ & $-58.7$ & N/A & N/A \\
\addlinespace[0.5ex]
\midrule
\textbf{Rewiring} & \multicolumn{6}{c}{\textbf{Node Classification Datasets}} \\
\cmidrule(lr){2-7}
 & \textbf{Cora} & \textbf{Citeseer} & \textbf{Texas} & \textbf{Cornell} & \textbf{Wisconsin} & \textbf{Chameleon} \\
\midrule
FoSR & $100.0$ & $150.0$ & $100.0$ & $100.0$ & $100.0$ & $37.5$ \\
DIGL & $154421.0$ & $96241.0$ & $2641.0$ & $5578.0$ & $30992.0$ & $41468.0$ \\
SDRF & $3.0$ & $0.0$ & $7.0$ & $0.0$ & $87.5$ & $68.5$ \\
BORF\tnote{*} & $0.0$ & $0.0$ & $16.5$ & $-12.5$ & $12.5$ & $-10.0$ \\
\bottomrule
\end{tabularx}
\begin{tablenotes}[flushleft]
\footnotesize
\item[*] For BORF, 0 or negative values are possible because this method may also remove edges. A negative value indicates that, on average, more edges were removed than added.
\end{tablenotes}
\end{threeparttable}
\end{table*}

% \section{ITE Scatter Plots}
% \label{app:ITEscatter}
% To examine the effect of rewiring strategies (Q1) on over-squashing in node classification tasks, we provide scatter plots of $\text{ITE}$ metrics values.

\end{document}